\documentclass[12pt, onecolumn, draftclsnofoot]{IEEEtran}
\usepackage{amsmath}
\usepackage{amssymb}
\usepackage{amsfonts}

\newcommand{\Rmnum}[1]{\expandafter\@slowromancap\romannumeral #1@}
\usepackage{graphicx}
\usepackage{epsfig}
\usepackage{subfigure}
\usepackage[sort]{cite}
\usepackage{xcolor}
\usepackage{enumerate}
\usepackage{extarrows}
\usepackage{algorithmic}
\usepackage{subfigure}
\usepackage[lined,ruled,linesnumbered]{algorithm2e}
\usepackage{tensor}
\usepackage[colorlinks,linkcolor=blue,anchorcolor=blue,citecolor=blue,urlcolor=black]{hyperref}

\begin{document}
\title{Pre-Training and Personalized Fine-Tuning via Over-the-Air Federated Meta-Learning: Convergence-Generalization Trade-Offs}

\author{Haifeng Wen, Hong Xing, and Osvaldo Simeone
\thanks{H. Wen and H. Xing are with the IoT Thrust, The Hong Kong University of Science and Technology (Guangzhou), Guangzhou, 511453, China; H. Xing is also affiliated with the Department of ECE, The Hong Kong University of Science and Technology, HK SAR (e-mails: hwen904@connect.hkust-gz.edu.cn,~hongxing@ust.hk). O. Simeone is with the Department of Engineering, King's College London, London, WC2R 2LS, U.K. (e-mail: osvaldo.simeone@kcl.ac.uk).}
}
\maketitle

\begin{abstract}
For modern artificial intelligence (AI) applications such as large language models (LLMs), the training paradigm has recently shifted to pre-training followed by fine-tuning. 
Furthermore, owing to dwindling open repositories of data and thanks to efforts to democratize access to AI models, pre-training is expected to increasingly migrate from the current centralized deployments to federated learning (FL) implementations. 
Meta-learning provides a general framework in which pre-training and fine-tuning can be formalized. Meta-learning-based personalized  FL (meta-pFL) moves beyond basic personalization by targeting generalization to new agents and tasks. 
This paper studies the generalization performance of meta-pFL for a wireless setting in which the agents participating in the pre-training phase, i.e., meta-learning, are connected via a shared wireless channel to the server. 
Adopting over-the-air computing, we study the trade-off between generalization to new agents and tasks, on the one hand, and convergence, on the other hand. 
The trade-off arises from the fact that channel impairments may enhance generalization, while degrading convergence. Extensive numerical results validate the theory.
\end{abstract}
%

\IEEEpeerreviewmaketitle
\newtheorem{definition}{\underline{Definition}}[section]
\newtheorem{fact}{Fact}
\newtheorem{assumption}{Assumption}
\newtheorem{theorem}{\underline{Theorem}}[section]
\newtheorem{lemma}{\underline{Lemma}}[section]
\newtheorem{proposition}{\underline{Proposition}}[section]
\newtheorem{corollary}[proposition]{\underline{Corollary}}
\newtheorem{example}{\underline{Example}}[section]
\newtheorem{remark}{\underline{Remark}}[section]
\newcommand{\mv}[1]{\mbox{\boldmath{$ #1 $}}}
\newcommand{\mb}[1]{\mathbb{#1}}
\newcommand{\Myfrac}[2]{\ensuremath{#1\mathord{\left/\right.\kern-\nulldelimiterspace}#2}}
\newcommand\Perms[2]{\tensor[^{#2}]P{_{#1}}}

\section{Introduction}

For modern artificial intelligence (AI) applications such as large language models (LLMs), the training paradigm has shifted to \emph{pre-training}  followed by \emph{fine-tuning}{\color{black} \cite{touvron2023llama}}. Pre-training is currently done centrally using large data repositories obtained, often at a high cost, by large corporations{\color{black} \cite{achiam2023gpt, touvron2023llama, team2023gemini}}; while fine-tuning is typically much cheaper, and it can be used to \emph{personalize} models to individual requirements{\color{black} \cite{hu2021lora,wei2021finetuned}}. For example, a foundation LLM can be fine-tuned to serve as a personal assistant or as a surrogate CEO based on data tailored to the individual use cases{\color{black}\cite{wei2021finetuned}}.
As large dataset repositories are becoming a scarce commodity, it has become imperative to use distributed data in a privacy-minded way for pre-training and, possibly, fine-tuning \cite{sun2025federated, ni2025federated}. 
Furthermore, efforts to democratize access to AI models are moving in the direction of federated pre-training implementations, which leverage decentralized computing resources~\cite{mcmahan2017communication}. 

\begin{figure}
    \centering
    \includegraphics[width=0.8\linewidth]{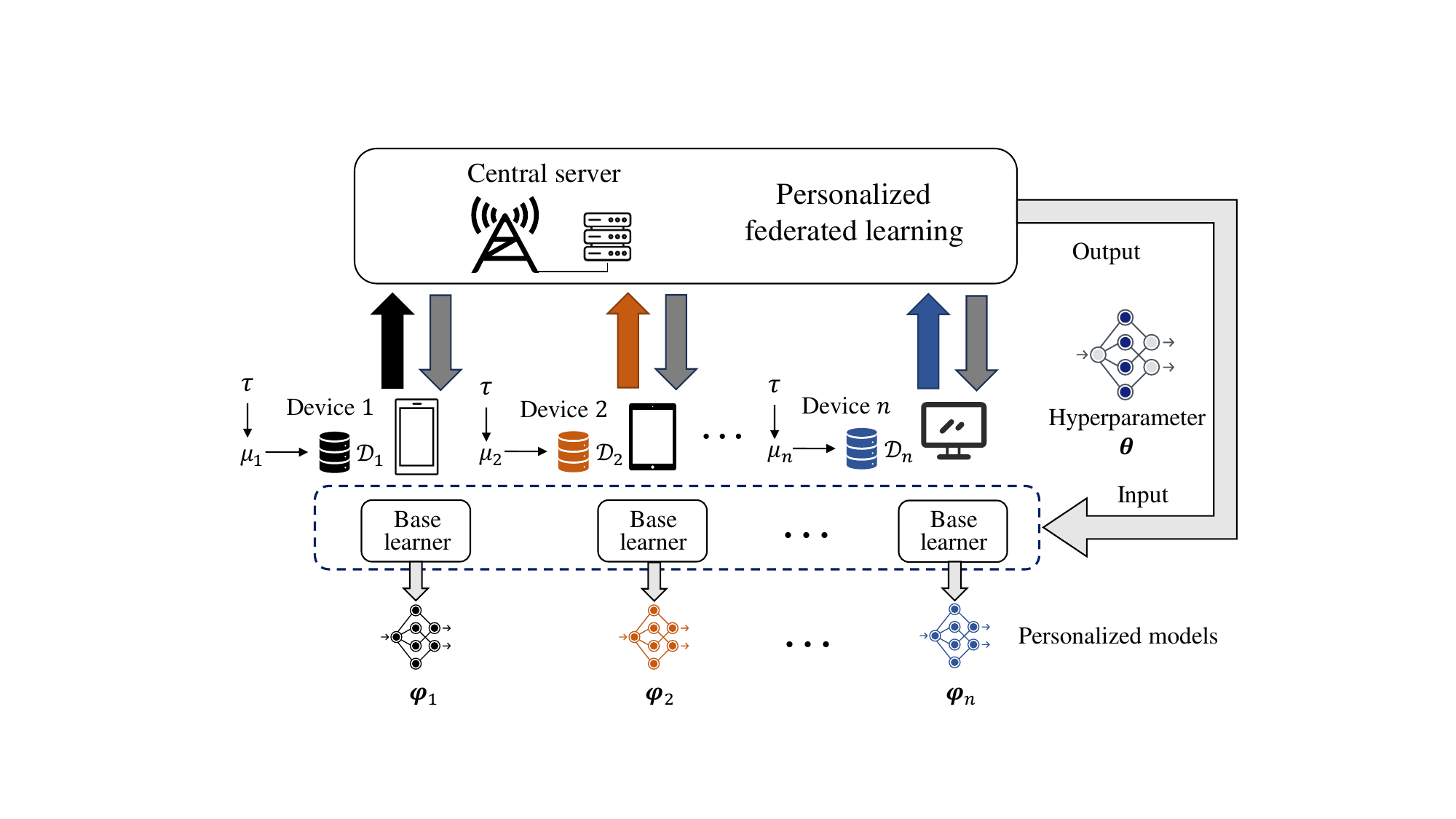}
    \vspace{-0.2in}
    \caption{In the considered personalized federated learning (pFL) setting, each device $i$ aims to find a {\color{black}fine-tuned} model $\mv \varphi_i\in \mathbb{R}^{d}$ using a local data set $\mathcal{D}_i\sim \mu_i^m$ by communicating with the central server. The central server maintains a hyperparameter $\mv \theta \in \mathbb{R}^{d}$, representing a pre-trained model, which is updated based on information {\color{black}sent by} the devices.} 
    \label{fig:pFL_illustration}
\end{figure}

A general framework in which pre-training and fine-tuning can be formalized is \emph{meta-learning}. In meta-learning, data from different tasks are used to pre-train a model with the aim of ensuring that the pre-trained model can be efficiently fine-tuned based on limited data for a new task \cite{chen2023learning,simeone2022machine}. 
When applied to a multi-agent, and federated setting, as illustrated in Fig. \ref{fig:pFL_illustration},  agents collaborate via communication to a central server for pre-training of a shared model defined by hyperparameters $\mv \theta$, from which each agent $i$ can extract a fine-tuned model $\mv \varphi_i$  that is tailored to the individual data sets at each agent $i$. As in federated learning (FL), agents do not directly exchange data but only model information, thus limiting the leakage of private information \cite{mcmahan2017communication}.

Conventional FL, which aims at finding a single shared model for all agents, is known to suffer from the heterogeneity of the distributions across agents. Heterogeneity has been addressed via methods such as{ decomposition~\cite{arivazhagan2019federated}, masking~\cite{setayesh2022perfedmask}, and clustering~\cite{mansour2020three}.} \emph{Personalized FL} (pFL)  alleviates the problem of data heterogeneity by optimizing personalized AI models that are adjusted to the local tasks of interest for each user~\cite{Tan2022towards}. 

\emph{Meta-learning-based pFL} (meta-pFL) moves beyond basic personalization by targeting \emph{generalization to new agents and tasks} \cite{fallah2020personalized, wang2023memory, khodak2019adaptive}. In line with the goal of the pre-training/fine-tuning workflow, meta-pFL does not merely aim for enhancing the performance of existing agents and tasks. Rather, it optimizes shared hyperparameters $\mv \theta$ that can be efficiently personalized via fine-tuning to \emph{new} agents and tasks that are not observed during federated pre-training.

This paper studies the generalization performance of meta-pFL for a wireless setting in which the agents participating in the pre-training phase, i.e., meta-learning, are connected via a shared wireless channel to the server~\cite{yang2020federated,zhu2019broadband}. 
{\color{black}
Wireless meta-pFL systems are suitable for large-scale intelligent Internet of Things (IoT) systems, in which IoT devices collect data for collaborative model training.
However, the performance of such systems may suffer from limited access to wireless resources.
}
A popular class of methods has explored the use of over-the-air computing \cite{nazer2007computation}, an approach known as \emph{over-the-air FL} (AirFL)~\cite{nazer2007computation,yang2020federated,zhu2019broadband}, to overcome communication bottlenecks on the shared radio channel for conventional FL. 

Channel impairments may not necessarily be deleterious for generalization. In fact, the addition of noise during pre-training can have a regularizing effect{\color{black} \cite{russo2019much}}, and it may even implement approximate forms of Bayesian learning \cite{simeone2022machine}{\color{black} \cite{welling2011bayesian}}. Accordingly, references such as{\color{black} \cite{liu20privacy, liu2022wireless, yang2021revisiting}} have shown improved generalization performance as a result of transmissions of model information over noisy channels in AirFL. The downside of a noisy training process is the convergence time, which may be severely affected by noisy updates{\color{black} \cite{sery2020analog, sery2021over, wen2023convergence}}. This paper studies the trade-off between generalization to new agents and tasks, on the one hand, and convergence, on the other hand, for an over-the-air implementation of meta-pFL.

\subsection{Related Works}
AirFL protocols for conventional FL have been widely studied. Among some notable algorithmic contributions,  the authors in \cite{zhu2019broadband,WCNC2023fullversion} proposed and optimized truncated-channel-inversion power control strategies, while reference \cite{amiri2020federated} proposed to enhance spectral efficiency via sparsification and linear compression mechanisms. Convergence was studied in ~\cite{sery2020analog,sery2021over} for phase compensation-based or channel inversion-based AirFL schemes targeting convex loss functions, showing a linear convergence rate as in the centralized gradient descent (GD). These studies reveal the impact of channel fading, channel noise, and data heterogeneity on the convergence performance of AirFL. Other related works on FL include \cite{cao2021optimized,wen2023convergence}, which derived and optimized transmission powers to minimize a convergence error upper bound of AirFL. 
 
Several pFL protocols have been proposed that apply in the presence of ideal channels. For instance, some methods leverage knowledge distillation (KD) to obtain personalized models by transferring knowledge of a powerful model to local models  \cite{li2019fedmd}; while others divide the trainable model parameters into personalized and shared parts \cite{arivazhagan2019federated}. The meta-pFL protocol studied in \cite{fallah2020personalized} builds on the model agnostic meta-learning paradigm (MAML) \textcolor{black}{\cite{finn2017model, rajeswaran2019meta}}, which applies a two-level gradient descent approach to learn how to quickly adapt to new tasks. Reference \cite{fallah2020personalized} analyzed the convergence of meta-pFL in the presence of ideal communication. 

Limited work has been done on the design of pFL protocols for wireless systems. Solutions include \cite{yue2022efficient,you2023automated}, cluster-based pFL \cite{li2022energy, zhao2023ensemble} and partial-model-based pFL \cite{chen2023knowledge}, which are based on digital transmission. The analysis in these papers provides insights into convergence. AirFL-based schemes for pFL were considered in \cite{chen2023personalizing,mortaheb2022personalized, sami2022over}, focusing on convergence analysis and optimal resource allocation.

For meta-pFL, the \emph{generalization error} is a key performance metric, as it quantifies the capacity of a shared model to be efficiently fine-tuned to new tasks.
 Generalization analyses for conventional AirFL can be found in~\cite{yang2021revisiting}, which shows that heavy-tail noise deteriorates the convergence rate while improving the generalization capacity. 
References \cite{jose2021information, chen2021generalization} have studied the generalization error for {MAML} under the assumption of ideal channels. 
To the best of our knowledge, no analysis of generalization exists for {meta-pFL under the assumption of ideal channels, let alone noisy channels.}

\subsection{Contributions}
This paper investigates a wireless implementation of the meta-pFL scheme that leverages over-the-air computing, with the main goal of understanding the trade-off between convergence and generalization entailed by the wireless transmission of model information. 
The main contributions are as follows:
\begin{enumerate}
    \item We introduce a  MAML-based over-the-air meta-pFL protocol, which adapts the approach in \cite{fallah2020personalized} for use of shared wireless channels. The proposed protocol, termed \emph{Air-meta-pFL}, leverages sparsification and linear compression{\color{black} \cite{amiri2020federated}} along with a long-term memory mechanism to compensate for the error caused by the gradient sparsification{\color{black} \cite{stich2018sparsified}}. 
    \item Convergence bounds are derived for  Air-meta-pFL under general smooth and non-convex loss functions with constant and adaptive learning rates. Our results quantify the impact of data heterogeneity, number of active devices, transmit power, number of channel uses, channel fading, and channel noise. 
    \item An upper bound on the generalization error of Air-meta-pFL is obtained that depends on the mutual information between the model parameters and the training data sets. The derived bound captures the impact of the same factors affecting convergence, namely data heterogeneity, number of active devices, transmit power, number of channel uses, channel fading, and channel noise. As anticipated, the analysis reveals a trade-off between convergence and generalization, with factors enhancing generalization, such as channel noise, potentially impairing convergence.
    \item Experimental results corroborate the insights gained from the convergence and generalization bounds, demonstrating the convergence-generalization trade-offs in practical conditions.
\end{enumerate}

The paper is organized as follows. The system model in Section \uppercase\expandafter{\romannumeral2} introduces the original pFL protocol and the communication model. Section \uppercase\expandafter{\romannumeral3} presents the proposed Air-meta-pFL protocol. Section \uppercase\expandafter{\romannumeral4} and \uppercase\expandafter{\romannumeral5} describe the derived upper bounds on convergence and generalization error, respectively. Numerical results are obtained in \uppercase\expandafter{\romannumeral6}. Finally, we conclude the paper in \uppercase\expandafter{\romannumeral7}.

\vspace{2ex}
\section{\textcolor{black}{Preliminaries and System Model}}\label{sec:System Model and Problem Statement}
In this section, we first review pFL and meta-pFL \cite{fallah2020personalized}, which operate under the assumption of ideal and noiseless communications. Then, we describe the considered wireless channel model that accounts for fading and noise. Throughout this work, we study the setting in Fig.~\ref{fig:pFL_illustration}, in which a set  $[n] \triangleq \{1, \ldots, n\}$ of devices communicate with an edge server over a multiple access (MAC) fading channel.  

\subsection{Preliminaries} \label{sec:Learning Protocol}
\textcolor{black}{\textbf{Personalized Federated Learning (pFL):}} As illustrated in Fig. \ref{fig:pFL_illustration}, each device $i \in [n]$ is assumed to possess a distinct local data set $\mathcal{D}_{i}=\{Z_{i,j}\}_{j\in [m]}$, where data point $Z_{i,j}$ with $j\in[m]$ is independent and identically distributed ($i.i.d.$) over a sample space $\mathcal{Z}$. The unknown data-generation distribution for device $i$ is denoted as $\mu_i$, so that each $j$-th data point in data set $\mathcal{D}_i$ is distributed as $Z_{i,j} \sim \mu_i$, and the overall local data set $\mathcal{D}_i$ is distributed as $\mathcal{D}_{i} \sim \mu_i^{m}$.
The data distributions $\{\mu_i\}_{i\in [n]}$ of all $n$ devices are drawn $i.i.d.$ from a common distribution $\tau$, i.e., $\mu_i \sim \tau$ for $i\in[n]$. Distribution $\tau$ captures the similarity of the data distribution across devices.

In pFL, the goal of each device $i$ is to minimize the \emph{local test loss}
\begin{equation}
    f_i(\mv \varphi_i) = \mathbb{E}_{Z\sim \mu_i}[\ell(\mv \varphi_i; Z)] \label{eq:local loss function}
\end{equation}
over a parameter vector $\mv \varphi_i \in \mathbb{R}^{d}$, where $\ell:\mathbb{R}^{d} \times \mathcal{Z} \to \mathbb{R}$ is a loss function. 
To this end, using a common vector $\mv \theta$, representing the \emph{pre-trained} model,   each device $i\in[n]$ applies a \emph{base learning algorithm} $\mathrm{P}_{\boldsymbol{\varphi}_i | \boldsymbol{\theta},\mathcal{D}_i}$, which implements a, generally stochastic, mapping between the local data set \(\mathcal{D}_i\) and the \textcolor{black}{pre-trained vector} $\mv \theta$ to a \textcolor{black}{fine-tuned}  model parameter $\mv \varphi_i$.
We denote by  
\begin{equation} \label{eq:local meta-test loss}
    F_i(\mv \theta) = \mathbb{E}_{\mathcal{D}_i\sim \mu_i^m}\left[\mathbb{E}_{{\boldsymbol{\varphi}}_i\sim \mathrm{P}_{\boldsymbol{\varphi}_i\mid \boldsymbol{\theta}, \mathcal{D}_i}} \left[f_i({\mv \varphi_i})\right]\right]
\end{equation}
the \emph{local meta-test loss} as a function of the \textcolor{black}{pre-trained vector} $\mv \theta$. Note that the outer expectation is with respect to (w.r.t.) the distribution $\mu_i^m$ of the local data set $\mathcal{D}_i$, and the inner expectation is over the output of the base learning algorithm.

The \textcolor{black}{pre-trained vector} $\mv \theta$ is ideally selected so as to address the problem of minimizing the \emph{meta-test loss} $F(\mv \theta)$, which is the average across all devices of the local meta-test loss \eqref{eq:local meta-test loss}. The corresponding optimization problem is given by 
\begin{equation} \label{eq:population risk minimization}
        \mathop{\mathtt{Minimize}}_{\boldsymbol \theta\in \mathbb{R}^{d}}~~~ \left\{ F(\mv \theta) = \frac{1}{n}\sum_{i=1}^n F_i(\mv \theta) \right\}.
\end{equation}

In order to address problem \eqref{eq:population risk minimization}, each device $i\in[n]$ divides the local data set $\mathcal{D}_i$ into two disjoint subsets $\mathcal{D}_i^{\rm (tr)}$ and $\mathcal{D}_i^{\rm (va)}$ such that $\mathcal{D}_i=\mathcal{D}_i^{\rm (tr)} \bigcup \mathcal{D}_i^{\rm (va)}$ with $|\mathcal{D}_i^{\rm (tr)}|=m^{\rm (tr)}$ and $|\mathcal{D}_i^{\rm (va)}|=m^{\rm (va)}$. 
The training data set $\mathcal{D}_i^{\rm (tr)}$ is utilized to implement the \emph{base learner} $\mathrm{P}_{\boldsymbol{\varphi}_i | \boldsymbol{\theta}, \mathcal{D}_i^{\rm (tr)} }$, while the validation data set $\mathcal{D}_{i}^{\rm (va)}$ is used to estimate the local test loss \eqref{eq:local loss function}. In particular, the base learning algorithm \textcolor{black}{fine-tunes via} one or more steps of (stochastic) gradient descent (GD) the \textcolor{black}{\emph{shared pre-trained vector}} $\mv \theta \in \mathbb{R}^{d}$ using the training data set $\mathcal{D}_{i}^{(\rm tr)}$. Considering, for simplicity of illustration, a single GD step, the base learner outputs the \textcolor{black}{fine-tuned} model
\begin{equation} \label{eq:local inner update}
    \mv \varphi_i = \boldsymbol{\theta} - \frac{\alpha }{m^{(\rm tr)}} \sum_{Z \in \mathcal{D}_i^{(\rm tr)}} \nabla \ell(\mv \theta; Z),
\end{equation}
where $\alpha>0$ is the step size.
Using the validation data set $\mathcal{D}_i^{\rm (va)}$ to estimate the local test loss \eqref{eq:local loss function} yields the following empirical version of problem $\eqref{eq:population risk minimization}$,
\begin{equation} \label{eq:definition of R_D}
    \begin{aligned}
        \mathop{\mathtt{Minimize}}_{\boldsymbol \theta\in \mathbb{R}^{d}}~ \Bigg\{ \hat{F}_{\mathcal{D}_{1:n}}(\mv \theta) = \frac{1}{n}\sum_{i=1}^{n}\frac{1}{m^{(\rm{va})}}\sum_{Z^\prime \in \mathcal{D}_i^{(\rm{va})}} \ell\Bigg(\boldsymbol{\theta} - \frac{\alpha }{m^{(\rm tr)}} \sum_{Z \in \mathcal{D}_i^{(\rm tr)}} \nabla \ell(\mv \theta; Z); Z^\prime \Bigg) \Bigg\},
    \end{aligned}
\end{equation}
where $\mathcal{D}_{1:n} = \bigcup_{i=1}^{n}\mathcal{D}_i$. The objective $\hat{F}_{\mathcal{D}_{1:n}}(\mv \theta)$ in problem \eqref{eq:definition of R_D} is known as the \emph{meta-training loss}.

\textcolor{black}{\textbf{Meta-pFL:}}
\textcolor{black}{Meta-pFL} addresses problem \eqref{eq:definition of R_D} using gradient descent on the \textcolor{black}{shared hyperparameter}  $\mv \theta$. 
Specifically, at each communication round $t \in \{0, ..., T-1\}$, a fraction $rn$, with $r\in (0, 1]$, of devices are chosen uniformly at random to transmit to the server.
We denote as $\mathcal{I}^{(t)}\subseteq [n]$ the subset of devices \emph{active} at round $t$, which has cardinality $|\mathcal{I}^{(t)}|=rn$.
After receiving the current vector $\mv \theta^{(t)}$ at round $t$ from the server, each 
active device $i\in \mathcal{I}^{(t)}$ initializes  the hyperparameter vector $\mv \theta$ as $\mv \theta_{i}^{(t,0)} = \mv \theta^{(t)}$. Then, it performs $Q$ local SGD steps as
\begin{align}
\mv \theta_{i}^{(t, q+1)} = \mv \theta_{i}^{(t, q)}-\eta^{(t)}\hat\nabla F_{i}(\mv \theta_{i}^{(t, q)}), \label{eq:local updates}
\end{align} 
where $q \in\{0,\ldots, Q-1\}$ is the local-SGD index; $\eta^{(t)}$ denotes the step size at round $t$; and $\hat\nabla F_{i}(\mv \theta_i^{(t,q)})$ is an estimate of the true gradient $\nabla F_{i}(\mv \theta_i^{(t,q)})$ of the local meta-test loss \eqref{eq:local meta-test loss}. This estimate is obtained using mini-batches of device $i$'s local data set $\mathcal{D}_i$ as
\begin{equation}
    \begin{split}
        \hat{\nabla} F_{i}(\mv \theta_{i}^{(t, q)}) = \left(\mv I_d - \alpha \hat{\nabla}^2 f_{i}(\mv \theta_{i}^{(t, q)})\right)\hat{\nabla} f_i\left(\mv \theta_{i}^{(t, q)} - \alpha \hat{\nabla} f_{i}(\mv \theta_{i}^{(t, q)})\right), 
    \end{split}
    \label{eq:F_SGD} 
\end{equation}
where 
$
\hat{\nabla} f_{i}(\mv \theta_{i}^{(t, q)})=\frac{1}{|\mathcal{B}_{i}^{(t)}|} \sum_{Z\in \mathcal{B}_{i}^{(t)}} \nabla\ell(\mv \theta_{i}^{(t, q)}; Z)  
$
is an estimate of the gradient of the local test loss~\eqref{eq:local loss function} obtained from a mini-batch $\mathcal{B}_{i}^{(t)}\subseteq\mathcal{D}_i^{\rm (tr)}$ of data points; and the vector \(\hat{\nabla} f_i(\mv \theta_{i}^{(t, q)} - \alpha \hat{\nabla} f_{i}(\mv \theta_{i}^{(t, q)}))\) and Hessian matrix \(\hat{\nabla}^2 f_{i}(\mv \theta_{i}^{(t, q)})\) are similarly obtained from distinct mini-batches of $\mathcal{D}_i^{\rm (va)}$. Note that the estimate $\hat{\nabla} F_{i}(\mv \theta)$ is not unbiased, i.e., $\mathbb{E}[\hat{\nabla} F_{i}(\mv \theta)] \ne \nabla F_{i}(\mv \theta)$~\cite{fallah2020personalized}. 

Each device $i \in \mathcal{I}^{(t)}$ transmits the hyperparameter model difference 
\begin{equation} \label{eq:model difference}
    \mv{\Delta}_{i}^{(t)} = \mv \theta_{i}^{(t,0)}-\mv \theta_{i}^{(t, Q)}
\end{equation}
to the server. The server aggregates the model differences to update the global hyperparameter vector $\mv \theta^{(t+1)}$ as
\begin{equation}
    \mv \theta^{(t+1)} = \mv \theta^{(t)} - \frac{1}{rn}\sum_{i\in \mathcal{I}^{(t)}}\mv{\Delta}_{i}^{(t)}. \label{eq:global updates}
\end{equation} 
The updated $\mv\theta^{(t+1)}$ is then broadcast to all $n$ devices at the beginning of round $t+1$. The above steps are iterated until a convergence criterion is met.

\subsection{Communication Model}\label{sec:Communication Model}
The \textcolor{black}{meta-pFL} protocol assumes ideal communication between devices and the server. In this subsection, we present a standard over-the-air computing (AirComp) communication model that accounts for limitations imposed by wireless multiple access channels. 

At the $t$-th communication round, it takes place over a block of $M$ channel uses. 
Each active device $i\in \mathcal{I}^{(t)}$ transmits an $M$-dimensional vector ${\mv{x}}_i^{(t)}$. The average transmit power $\mathbb{E}||{\mv{x}}_i^{(t)}||^2$ must satisfy the constraint as
\begin{equation} \label{eq:power constraint}
    \frac{1}{M} \mathbb{E}\| \boldsymbol x_{i}^{(t)} \|^2 \le P_{i},
\end{equation}
where the expectation is over the transmitted signal $\mv x_i^{(t)}$, and $\|\cdot\|$ is denoted as $l_2$-norm in vector space.
We assume that complex channel $h_{i}^{(t)}=|h_{i}^{(t)}|e^{j\phi_{i}^{(t)}}$ between each device $i\in [n]$ and the server remains constant for the duration of a block, but may vary from block to block.
All active devices transmit simultaneously over the MAC, so that the edge server receives an $M\times 1$ vector given by
\begin{equation}
    \mv y^{(t)}=\sum_{i\in \mathcal{I}^{(t)}}h_{i}^{(t)}\mv x_{i}^{(t)}+\mv n^{(t)},
    \label{eq:rxsignal}
\end{equation}
where $\mv n^{(t)}$ is the additive white Gaussian noise (AWGN) vector with zero mean and variance $\sigma_n^2$.
{\color{black}
We also assume that the interference from non-AirFL IoT devices is negligible or can be eliminated via signal processing methods \cite{jha2023analog}.
}

As in \cite{amiri2020federated}, we assume ideal and noiseless communication in the downlink, given the less constrained resources available for downlink communication at the edge server. 

\section{\textcolor{black}{Over-the-Air Meta-Learning Based Personalized Federated Learning}}
\begin{figure}[t]
    \centering
    \includegraphics[width=1\linewidth]{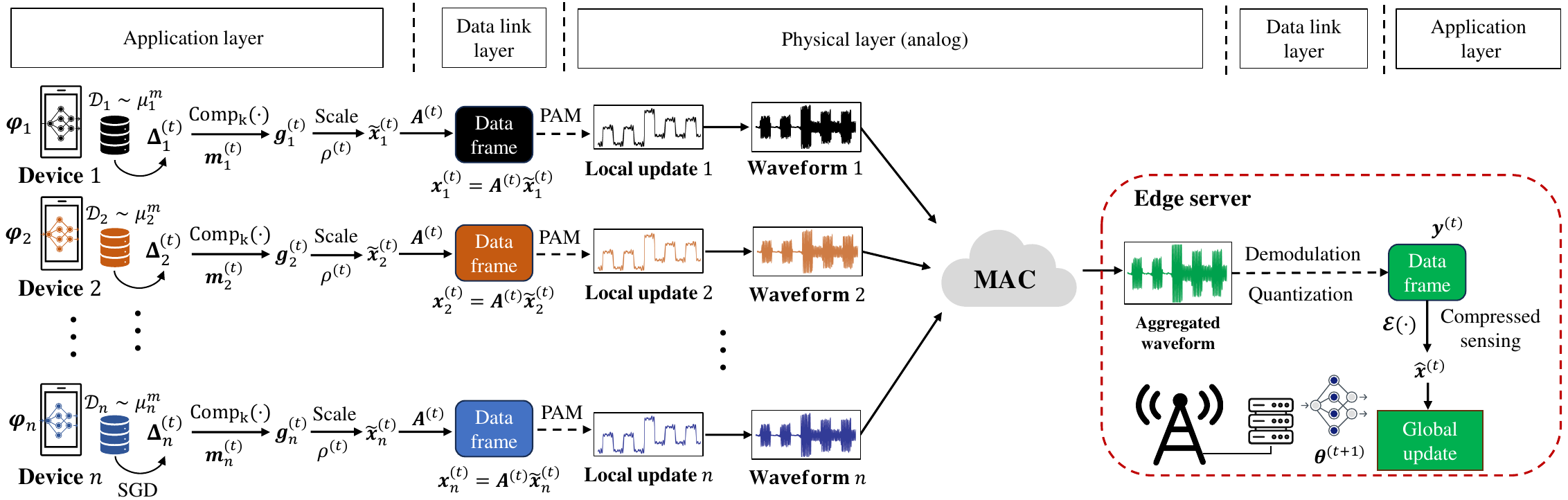}
    \vspace{-0.4in}
    \caption{\textcolor{black}{Illustration of the implementation protocol of over-the-air meta-pFL.}}
    \label{fig:illustration_AirFL}
\end{figure}

\textcolor{black}{In this section, we describe the proposed implementation protocol of over-the-air meta-pFL as illustrated in Fig.~\ref{fig:illustration_AirFL}.}
The analysis of convergence and generalization of the approach will be provided in the next two sections.

\subsection{\textcolor{black}{Air-meta-pFL}}\label{subsec:Air-meta-pFL}
As described in the previous section, in \textcolor{black}{meta-pFL} at each communication round $t$, each active device $i\in \mathcal{I}^{(t)}$ must communicate its local update $\mv \Delta_i^{(t)}$ in \eqref{eq:model difference} to the server.
To this end, as in \cite{amiri2020federated, sery2020analog, xing2021federated, wen2023convergence}, we adopt an analog communication strategy based on sparsification, linear compression, channel phase compensation, and power scaling.

First, a $k$-contraction operator is applied to sparsify vector $\mv \Delta_i^{(t)}$.
The $k$-contraction operator $\text{Comp}_k:\mathbb{R}^{d} \to \mathbb{R}^{d}$ preserves only $k$ entries of its input vector, while setting all other entries to zeros, e.g., via a top-$k$ selection or via a random selection strategy \cite{stich2018sparsified}. 
The resulting sparsified update is given by
\begin{equation} \label{eq:Comp-k}
\mv{g}^{(t)}_i=\text{Comp}_k(\mv{m}_{i}^{(t)}+\mv{\Delta}_{i}^{(t)}),
\end{equation}
where a memory vector $\mv{m}_{i}^{(t)}\in\mb{R}^{d}$ is used to keep track of the accumulated errors as \cite{stich2018sparsified}
\begin{equation} \label{eq:memory update}
    \mv{m}_{i}^{(t+1)}= \mv{m}_{i}^{(t)}+ \left(\mv{\Delta}_{i}^{(t)} - \mv g_i^{(t)}\right).
\end{equation}
The sparsified hyperparameter model difference is scaled as
\begin{equation} \label{eq:scaled sparsified vector}
    \tilde{\mv x}_i^{(t)} = \frac{\sqrt{\rho^{(t)}}e^{-j\phi_{i}^{(t)}}}{\eta^{(t)}} \mv g_i^{(t)},
\end{equation}
where the factor $e^{-j\phi_{i}^{(t)}}$ is included to compensate for the phase of channel coefficient $h_{i}^{(t)}=|h_{i}^{(t)}|e^{j\phi_{i}^{(t)}}$, and the common scaling factor $\rho^{(t)}$ is chosen to meet the average power constraint \eqref{eq:power constraint}.

Finally, in order to reduce the dimensionality of the transmitted vector to fit the $M$ channel uses, each active device $i \in \mathcal{I}^{(t)}$ projects the scaled vector $\tilde{\mv x}_i^{(t)}$ using the same  $M\times d$ matrix $\mv A^{(t)}$ as
\begin{equation} \label{eq:Ax}
    \mv x_i^{(t)} = \mv A^{(t)} \tilde{\mv{x}}_i^{(t)}.
\end{equation} 
Matrix $\mv A^{(t)}$ is assumed to have the property that its spectral norm $\|\mv A^{(t)}\|_2$, i.e., the square root of the maximum eigenvalue of matrix $\mv A^{(t)}(\mv A^{(t)})^T$, satisfies the inequality $\|\mv A^{(t)}\|_2 \le 1$. This condition guarantees the power constraint \eqref{eq:power constraint}. Such a compression matrix can be generated by, e.g., selecting $M$ rows of any unitary matrix. 

By substituting equation \eqref{eq:Ax} in \eqref{eq:rxsignal}, the vector received at the server can be written as
\begin{equation} \label{eq:received signal}
    \mv{y}^{(t)}=\mv{A}^{(t)}\sum_{i\in\mathcal{I}^{(t)}} h_i^{(t)} \tilde{\mv{x}}_i^{(t)}+\mv{n}^{(t)}.
\end{equation}
The received signal $\mv{y}^{(t)}$ is used to produce an estimate  $\hat{\mv x}^{(t)} = \mathcal{E}(\mv{y}^{(t)})$ of the weighted sum of updates $\sum_{i\in\mathcal{I}^{(t)}} h_i^{(t)} \tilde{\mv{x}}_i^{(t)}$, which serves by \eqref{eq:Comp-k}-\eqref{eq:scaled sparsified vector} as an estimate of the sum of updates $\sum_{i\in\mathcal{I}^{(t)}} \mv \Delta_i^{(t)}$.
With such an estimate, the global hyperparameter is finally updated as
\begin{equation} \label{eq:over-the-air global updatas}
    \mv \theta^{(t+1)} = \mv \theta^{(t)} - \frac{\eta^{(t)}}{\mu_h \sqrt{\rho^{(t)}}rn}\hat{\mv x}^{(t)},
\end{equation}
approximating the ideal meta-pFL update \eqref{eq:global updates}, where $\mu_h = \mathbb{E}[|h_i^{(t)}|]$ for all $i$ and $t$ (c.f. Assumption~\ref{assumption:channel fading}). Note that the term $\Myfrac{\eta^{(t)}}{(\mu_h \sqrt{\rho^{(t)}}rn)}\hat{\mv x}^{(t)}$ is the unbiased estimate of $\Myfrac{1}{(rn)}\sum_{i \in \mathcal{I}^{(t)}}\mv g_i^{(t)}$.
The proposed Air-meta-pFL is summarized in Algorithm \ref{alg:Air-meta-pFL}.

\begin{algorithm}[!htp] \label{alg:Air-meta-pFL}
    \SetKwInOut{Input}{Input}
    \SetKwInOut{Output}{Output}
    \SetKwBlock{DeviceParallel}{Active devices $i \in \mathcal{I}^{(t)}$ (in parallel):}{end}
    \SetKwBlock{localSGD}{for $q=0$ to $Q-1$ do}{end}
    \SetKwBlock{OnServer}{Server:}{end}
    \caption{\textcolor{black}{Air-meta-pFL}}
    \textbf{Input:} Learning rate $\{\eta^{(t)}\}_{t=0}^{T-1}$ adaptation learning rate $\alpha$; sizes of minibatches $m_B$ for computing $\hat{\nabla} f_{i}(\mv \theta_{i}^{(t, q)})$,  \(\hat{\nabla} f_i(\mv \theta_{i}^{(t, q)} - \alpha \hat{\nabla} f_{i}(\mv \theta_{i}^{(t, q)}))\), and \(\hat{\nabla}^2 f_{i}(\mv \theta_{i}^{(t, q)})\) in \eqref{eq:F_SGD}; maximum transmit powers $\{P_i\}_{i\in[n]}$; sparsification rate $k/d$; number of local SGD steps $Q$; fraction $r\in(0, 1]$ of active devices; and estimator $\mathcal{E}(\cdot)$; \\ 
    Initialize $t=0$, $\mv \theta_i^{(0, 0)}=\mv \theta^{(0)}$, and $\mv m_i^{(0)} = \mv 0$, for all $i \in [n]$; \\
    \While{$t < T$}{
    \OnServer{
    Chooses a subset of $rn$ devices $\mathcal{I}^{(t)}$ uniformly at random; \\ 
    }
    \DeviceParallel{
    Initialize $\mv{\theta}^{(t,0)}_{i}\leftarrow \mv{\theta}^{(t)}$; \\ 
    Perform $Q$ local SGD steps via \eqref{eq:local updates} to obtain the local hyperparameter model difference $\mv{\Delta}_{i}^{(t)} = \mv \theta_i^{(t,0)} - \mv \theta_i^{(t,Q)}$; \\ 
    Sparsify: $\mv{g}^{(t)}_i=\text{Comp}_k(\mv{m}_{i}^{(t)}+\mv{\Delta}_{i}^{(t)})$ \;
    Memory update: $\mv{m}_{i}^{(t+1)}=\mv{m}_{i}^{(t)}+ \mv \Delta_i^{(t)} -\mv{g}_{i}^{(t)}$\;
    Scale: $\tilde{\mv x}_i^{(t)} = (\Myfrac{\sqrt{\rho^{(t)}}e^{-j\phi_{i}^{(t)}}}{\eta^{(t)}}) \mv g_i^{(t)}$\;
    Transmit $\mv x_i^{(t)} =\mv A^{(t)} \tilde{\mv x}_i^{(t)}$ to the server\;
    }
    \OnServer{
    Receive $\mv{y}^{(t)}=\mv{A}^{(t)}\sum_{i\in\mathcal{I}^{(t)}} h_i^{(t)} \tilde{\mv{x}}_i^{(t)}+\mv{n}^{(t)}$\;
    Estimate: $\hat{\mv x}^{(t)} = \mathcal{E}(\mv{y}^{(t)})$\;
    Global update: $\mv \theta^{(t+1)} \leftarrow \mv \theta^{(t)} - (\Myfrac{\eta^{(t)}}{(\mu_h \sqrt{\rho^{(t)}}rn}))\hat{\mv x}^{(t)}$\;
    Broadcast $\mv \theta^{(t+1)}$ to all the $n$ devices\;
    }
    $t \leftarrow t + 1$\;
    }
    \textbf{Output:} $\mv \theta^{(T)}$
\end{algorithm}

\section{Convergence Analysis}
In this section, we analyze the convergence of the proposed Air-meta-pFL method for general smooth and non-convex loss functions. 
The main goal is to understand the impact of key parameters such as the signal-to-noise ratio (SNR), the number of devices, data heterogeneity, and the size of the communication block on the number of communication rounds needed for convergence.
The next section will highlight the different roles that some of these parameters can play in terms of generalization.

\subsection{Assumptions} \label{subsec:assumptions}
Our analysis is based on the following assumptions.
\begin{assumption}[$k$-Contraction \cite{stich2018sparsified}] \label{assumption:contraction}
    For a parameter $0<k\le d$, the $k$-contraction operator used in \eqref{eq:Comp-k} satisfies the inequality
\begin{equation} \label{eq:k-contraction inequality}
    \mathbb{E}\|\mv{x}-\text{Comp}_k(\mv{x})\|^2 \le \left( 1 - \frac{k}{d} \right)\|\mv{x}\|^2.
\end{equation}
\end{assumption}

\begin{assumption}[Bounded Gradient's Norm, Variance and Hessian's Variance]~\label{assumption:Bounded Variance and Norm}
    For all $i\in[n]$, the gradient and the Hessian of the loss function $\ell(\mv \varphi; Z)$ have bounded variance, i.e.,
    \begin{equation} \label{eq:gradient bounded variance}
        \mb {E}_{Z \sim \mu_i}\left\| \nabla \ell (\mv \varphi ;Z) - \nabla f_i(\mv\varphi) \right\|^2 \le \sigma_G^2, \quad
        \mb {E}_{Z \sim \mu_i}\left\| \nabla^2 \ell (\mv \varphi;Z) - \nabla^2 f_i(\mv\varphi) \right\|_2^2 \le \sigma_H^2,
    \end{equation}
    with the norm of the gradient having also a bounded second moment given by
    \begin{equation} \label{eq:gradient bounded second moment}
        \mb {E}_{Z \sim \mu_i}\| \nabla \ell (\mv \varphi;Z) \|^2 \le G^2.
    \end{equation}
\end{assumption}

\begin{assumption}[Smoothness of the Local Test Loss] \label{assumption:gradient-Lipschitz}
    For all $i\in [n]$, the local test loss function \(f_i(\cdot)\) is continuously differentiable and $L$-smooth such that for any pair of vectors $ \mv \varphi$ and $\mv \varphi^{\prime} \in \mb{R}^{d}$, there exists a constant $L_{G}>0$ satisfying the inequality
    \begin{equation} \label{eq:L-smoothness}
        \| \nabla f_i(\mv \varphi) - \nabla f_i(\mv \varphi^{\prime}) \| \le L_{G} \|\mv \varphi - \mv \varphi^{\prime}\|.
    \end{equation}
    Furthermore, the Hessian of local test loss function \(f_i(\cdot)\) is $L_H$-Lipschitz continuous, i.e., for all pairs $\mv \varphi$ and $\mv \varphi^{\prime} \in \mathbb{R}^d$, we have the inequality
    \begin{equation} \label{eq:Hessian-Lipschitz}
        \| \nabla^2 f_i(\mv \varphi) - \nabla^2 f_i(\mv \varphi^{\prime})\|_2 \le L_H \| \mv \varphi - \mv \varphi^{\prime} \|
    \end{equation}
    for some $L_H>0$.
\end{assumption}

\begin{assumption}[Bounded Data Heterogeneity \cite{fallah2020personalized}] \label{assumption:Bounded Data Heterogeneity}
    For any $\mv \varphi \in \mathbb{R}^{d}$, the gradient $\nabla f_{i}(\mv \varphi)$ of the local test loss and its Hessian $\nabla^2 f_{i}(\mv \varphi)$ satisfy inequalities
    \begin{align} 
        \| \nabla f_{i}(\mv \varphi) - \nabla f(\mv \varphi) \|^2 \le \gamma_G^2, \quad
        \| \nabla^2 f_{i}(\mv \varphi) - \nabla^2 f(\mv \varphi) \|_2^2 \le \gamma_H^2, 
    \end{align}
    where $f(\mv \varphi)=\Myfrac{1}{n}\sum_{i=1}^{n}f_i(\mv \varphi)$.
\end{assumption}

\begin{assumption}[Estimation Error] \label{assumption:Estimation Error}
The estimate $\hat{\mv x}^{(t)}=\mathcal{E}(\mv y^{(t)})$ of the sum of the updates $\sum_{i\in \mathcal{I}^{(t)}}\mv \Delta^{(t)}$ can be expressed as  $\hat{\mv x}^{(t)} =\sum_{i\in\mathcal{I}^{(t)}} h_i^{(t)} \tilde{\mv{x}}_i^{(t)} + \mv{n}^{(t)}_{\text{est}}$, where the estimation error $\mv n_{\text{est}}^{(t)}$ has zero mean and variance 
$
    \mathbb{E}||\mv n_{\text{est}}^{(t)}||^2=d \cdot v^{(t)}    
$
for some $v^{(t)}>0$, and is uncorrelated with the signal $\sum_{i\in\mathcal{I}^{(t)}} h_i^{(t)} \tilde{\mv{x}}_i^{(t)}$. 
\end{assumption}

Note that Assumption~\ref{assumption:Estimation Error} is trivially satisfied for a scheme that does not implement sparsification and linear compression by setting the variance of the estimation error $v^{(t)}$ to be equal to the channel noise power, i.e., $ v^{(t)}= \sigma^2$ for all $t$. In the more general case, this assumption is satisfied by applying a linear minimum mean squared error (LMMSE) estimator. It is also approximately met by Bayesian sparse recovery algorithms \cite{ma2017orthogonal}, for which one can evaluate the estimation error $v^{(t)}$ by leveraging state evolution  \cite{donoho2009message, ma2017orthogonal}.  

\begin{assumption}[Time-Varying Channels] \label{assumption:channel fading}
    The channel coefficients $h_i^{(t)}$ are $i.i.d.$, with absolute value $|h_i^{(t)}|$ having finite mean $\mathbb{E}[|h_i^{(t)}|] = \mu_h$ and power $\mathbb{E}[|h_i^{(t)}|^2] = \sigma_h^2$.
\end{assumption}

\subsection{Convergence Analysis with Constant Learning Rates}
Under the assumptions listed above, the following convergence result holds for Air-meta-pFL with constant learning rates, i.e., with $\alpha^{(t)}=\alpha$ for the inner SGD update \eqref{eq:local inner update} and with $\eta^{(t)}=\eta$ for the outer SGD update \eqref{eq:local updates}. 
As we will see, fixed learning rates yield a convergence rate of the order $\mathcal{O}(1/\sqrt{T})$ in the number of communication rounds $T$, while exhibiting an error floor. The next subsection will show that the error floor can be eliminated via adaptive learning rates, but at the cost of a slower convergence rate.
\begin{theorem}[Convergence with Constant Learning Rate] \label{theorem:convergence}
    Under Assumptions~\ref{assumption:contraction}-\ref{assumption:channel fading}, let \(\{\mv\theta^{(t)}\}_{t=0}^{T-1}\) be the iterates generated by Air-meta-pFL (Algorithm~\ref{alg:Air-meta-pFL}) with constant learning rates $\alpha^{(t)} = \alpha \in (0, 1/L_G]$ and $\eta^{(t)} = \eta$ that satisfies the inequalities
    \begin{equation} \label{eq:requirement of learning rate}
        \begin{split}
            \eta \in \bigg\{\eta: ~ 0<\eta\le \frac{1}{10QL_F}, 60\eta^2Q^2L_F^2+160\eta^3Q^3L_F^3+4\eta QL_F \le \frac{1}{8} \bigg\}.
        \end{split}
    \end{equation}
    Then, on average over the randomness of SGD, sparsification, device selection, fading, and channel noise, Air-meta-pFL satisfies the following inequality
    \begin{equation} \label{eq:convergence bound}
        \frac{1}{T}\sum_{t=0}^{T-1}\mathbb{E}\| \nabla F(\mv \theta^{(t)}) \|^2 \le  \frac{C_0}{\eta T} + \eta(C_n + C_{v}) + \eta^2(C_{F}+C_{\Lambda}) + C_{\alpha},
    \end{equation}
    where
    {
    \begin{align*}
        C_0 & =  \underbrace{\frac{8(F(\mv \theta^{(0)}) - F^*)}{Q}}_{\text{Initialization error}}, \quad         C_{\alpha} = \underbrace{\frac{48 \alpha^2L^2\sigma_G^2}{m_B}}_{\text{Inner SGD error}}, \quad         C_{\Lambda} =   \underbrace{\frac{32Q^2 L_F^2 \Lambda}{r^2} \left( (1+\alpha L_{G})^2 + \frac{\alpha^2 \sigma_H^2}{m_B} \right)G^2}_{\text{Sparsification error}}, \\  
        C_{v} & = \underbrace{16L_F Q G^2(\Lambda + 1)\left( (1+\alpha L_{G})^2 + \frac{\alpha^2 \sigma_H^2}{m_B} \right)\left( \frac{d}{r^2n^2 MP_{\min}}\frac{1}{T}\sum_{t=0}^{T-1}v^{(t)} + \frac{2\sigma_h^2}{\mu_h^2} - 2 \right)}_{\text{Estimation error}} \\
        C_n & = \underbrace{ 32 Q L_F\left( \sigma_F^2 + \gamma_F^2 \right)}_{\text{Outer-SGD error \& data heterogeneity}}, \quad 
        C_F = \underbrace{(480 Q^2 L_F^2 + 1280 Q^3L_F^2)(\sigma_F^2+\gamma_F^2)}_{\text{Outer-SGD error \& data heterogeneity}}, 
    \end{align*}
    }
    where we have defined    
    $\lambda=k/d$, $L_F = 4L + \alpha L_H G$, $\gamma_F^2=3 G^2 \alpha^2 \gamma_H^2+192 \gamma_G^2$, $\Lambda = \frac{(1-\lambda)(1+1/c)}{1-(1-\lambda)(1+c)}$ with $0<c<\frac{\lambda}{1-\lambda}$, $\sigma_F^2=12B^2 \sigma_H^2\frac{\alpha^2}{4m_B}+12\sigma_G^2\left(\frac{1+(\alpha L_{G})^2}{m_B}\right)\left(1+\sigma_H^2 \frac{\alpha^2}{4 m_B}\right)$, and $P_{\min} = \min_{i\in [n]} P_i$.
\end{theorem}
{\color{black}
\begin{IEEEproof}
    The proof is provided in Appendix~\ref{appendix:proof of convergence}. 
 \end{IEEEproof}
 }

As anticipated, the bound in \eqref{eq:convergence bound} indicates a convergence to a bounded error with a rate $\mathcal{O}(1/\sqrt{T})$ by setting $\eta$ with the order $\mathcal{O}(1/\sqrt{T})$. The error floor $C_{\alpha}$ depends on the adaptation learning rate $\alpha$ used in the inner adaptation SGD update \eqref{eq:local inner update}, as well as on the variance $\sigma_G^2$ of the corresponding gradient.
At convergence, the performance is thus limited by the noise in the adaptation step, which can be reduced by increasing the corresponding mini-batch size $m_B$.

The convergence speed, determined by the term scaling as $1/\sqrt{T}$ is dictated by:
(\emph{i}) The term $C_0$, accounts for the initialization error; 
(\emph{ii}) The term $C_n$, which depends on the SGD error on the outer update \eqref{eq:local updates} for the hyperparameter vector, increases with the number $Q$ of outer update steps and with data heterogeneity via parameter $\gamma_F$;
and (\emph{iii}) by the term $C_v$, which depends on the variance of the estimation error $v^{(t)}$ due to channel noise, sparsification, and linear compression. This latter term decreases with the number of active devices $rn$, with the block size $M$, and with the SNR via the power $P_{\min}$, while increasing with the level of data heterogeneity. 

The bound \eqref{eq:convergence bound} also includes a faster contribution that scales as $\mathcal{O}(1/T)$ with $\eta$ set as the order $\mathcal{O}(1/\sqrt{T})$.
In a manner similar to the part decreasing as $\mathcal{O}(1/\sqrt{T})$, this contribution is determined by the SGD error on the hyperparameter update, on data heterogeneity, and on the sparsification error.
In particular, the term $C_F$ increases with the variance $\sigma_F^2$ of the outer-SGD updates, with the level of data heterogeneity $\gamma_F^2$, and with the number $Q$ of updates; 
while the term $C_{\Lambda}$ decreases with $k$, and with the number $Q$ of updates, while decreasing with the mini-batch size $m_B$.

\subsection{Convergence Analysis with Adaptive Learning Rates}
The following Theorem provides an asymptotic analysis of the convergence based on adaptive learning rates.
A suitable choice of decreasing learning rates can remove the error floor in \eqref{eq:convergence bound} due to the noise in the adaptation step, but at the cost of a slower learning rate.
\begin{theorem}[Convergence with Adaptive Learning Rate] \label{theorem:convergence adaptive}
    Under Assumptions~\ref{assumption:contraction}-\ref{assumption:channel fading}, let \(\{\mv\theta^{(t)}\}_{t=0}^{T-1}\) be iterates generated according to Air-meta-pFL (Algorithm~\ref{alg:Air-meta-pFL}) with an inner learning rate $\alpha^{(t)} = \Myfrac{\xi^\prime}{(a^\prime + t)} \in (0, 1/L_{G}]$ and an outer learning rate $\eta^{(t)}=\Myfrac{\xi}{(a+t)}$ that satisfies $0<\eta^{(0)}\le \Myfrac{1}{(10QL_F)}$ and $60(\eta^{(0)})^2Q^2L_F^2+160(\eta^{(0)})^3Q^3L_F^3+4\eta^{(0)}QL_F \le \Myfrac{1}{8} $
    Then, on average over the randomness of SGD, sparsification, device selection, fading, and channel noise, Air-meta-pFL satisfies the following inequality
    \begin{equation} \label{eq:asymptotic convergence bound}
        \min_{t=0,1,\ldots,T-1} \mathbb{E}\| \nabla F(\theta_t) \|^2 \le \frac{C_{\text{ada}}}{\xi\ln(\frac{T+a-1}{a})},
    \end{equation}
    where
    {
    \begin{multline} 
        C_{\text{ada}} = \underbrace{\frac{8 F(\mv \theta^{(0)})-F^* }{ Q}}_{\text{Initialization error}}+
        \underbrace{\frac{48 L_{G}\sigma_G^2}{ m_B}\frac{(\xi^\prime)^2}{a^\prime - 1}}_{\text{Inner SGD error}}
        + \underbrace{64 Q^2L_F^2G^2\frac{C}{r^2 \lambda^2}\left(1+\frac{\sigma_H^2}{L_{G}^2 m_B}\right)\frac{\xi^3}{(a-1)}}_{\text{Sparsification error}}  \\ 
        +\underbrace{{4L_F Q G^2 \left(1+\frac{\sigma_H^2}{L_{G}^2m_B}\right)\left(\frac{C}{\lambda^2}+2\right) }\frac{\xi^2}{(a-1)}\left( \frac{d}{r^2n^2 M P_{\min}} \max_t v^{(t)} + \frac{2\sigma_h^2}{\mu_h^2} - 2 \right)}_{\text{Estimation error}} \\   
        +\underbrace{\left(480 Q^2L_F^2 + 1280 Q^3L_F^2\right)(\sigma_F^2+\gamma_F^2)\frac{\xi^3}{(a-1)^2}}_{\text{Outer-SGD error \& data heterogeneity}} 
        + \underbrace{\frac{32 QL_F \xi^2 }{(a-1)}\left(  \sigma_F^2 + \gamma_F^2 \right)}_{\text{Outer-SGD error \& data heterogeneity}},
    \end{multline}
    }
    where $L_F$, $\gamma_F^2$, $\sigma_F^2$, and $P_{\min}$ are defined as in Theorem~\ref{theorem:convergence}, and $C$ is a constant satisfying the inequality $C\ge \frac{4a\lambda(1-\lambda^2)}{a\lambda - 4Q}$.
\end{theorem}
{\color{black}
\begin{IEEEproof}
The sketch proof is provided in Appendix~\ref{appendix:proof of convergence}.
 \end{IEEEproof}
}
 
This result indicates that Air-meta-pFL converges with rate $\mathcal{O}(1/\ln(T))$ when choosing adaptive learning rates as decaying $\mathcal{O}(1/t)$ over the communication round index $t$. 
The constant $C_{\text{ada}}$ in \eqref{eq:asymptotic convergence bound} depends on the same factors as the constants in \eqref{eq:convergence bound}. In particular, the term $C_{\text{ada}}$ decreases with the effective SNR level $P_{\min}/v^{(t)}$, the number of active devices $rn$, and the block size $M$, while increasing with the level of the data heterogeneity $\gamma_F^2$. 


\section{Generalization Analysis} \label{sec:generalization}
Air-meta-pFL aims at finding a \textcolor{black}{shared pre-trained vector} $\mv \theta$ that supports the design of effective \textcolor{black}{fine-tuned} models $\mv \varphi$ via an SGD-based learner at the device. 
The previous section took an optimization perspective, addressing the convergence of the protocol to stationary points of the meta-training loss minimization problem~\eqref{eq:definition of R_D}. 
However, convergence does not provide any guarantee in terms of test performance. In fact, improving the accuracy of a solution to problem~\eqref{eq:definition of R_D} may harm the test performance due to overfitting.
To provide a more complete assessment of the performance of Air-meta-pFL, in this section, we analyze the \emph{generalization} performance.
As we will argue, communication-related variables, such as received SNR, number of active devices, and number of channel uses, may play opposite roles when viewed through the lenses of convergence and of generalization. This analysis will highlight the emergence of a possible trade-off between convergence and generalization, which will be further studied in Sec. \ref{sec:Numerical Results} via numerical results.

\subsection{Meta-Generalization Error}
As in the generalization analysis of MAML \cite{chen2021generalization}, we study the generalization performance of Air-meta-pFL in terms of the local test loss experienced by a new device that did not participate in the federated learning process.
Recalling that each device is characterized by a data distribution $\mu \sim \tau$, let us write the corresponding meta-test loss \eqref{eq:local meta-test loss} for the new device directly as a function of the data-generation distribution $\mu$ as the expectation 
\begin{equation}
    F_{\mu}(\mv \theta) = \mathbb{E}_{\mathcal{D}\sim \mu}\mathbb{E}_{\boldsymbol{\varphi} \sim \mathrm{P}_{\boldsymbol{\varphi}|\boldsymbol{\theta}, \mathcal{D}}}\mathbb{E}_{Z\sim \mu}[\ell(\mv \varphi; Z)].
\end{equation}
Accordingly, the expected \emph{meta-test loss} for the new device is defined as 
\begin{equation} \label{eq:definition of R_tau}
    \bar{F}_{\tau}(\mv \theta) \triangleq \mathbb{E}_{\mu\sim \tau}[F_{\mu}(\mv \theta)],
\end{equation} 
which is averaged over the possible local distributions $\mu$. 

Generalization is measured via the expected gap between meta-training loss $\hat{F}_{\mathcal{D}_{1:n}}(\mv \theta)$ optimized by meta-pFL as per problem \eqref{eq:definition of R_D} and the target meta-test loss $\bar{F}_{\tau}(\mv \theta)$. If this gap is small, then convergence to a solution of problem \eqref{eq:definition of R_D} does indeed enforce a good generalization performance, while this is not the case otherwise. The resulting \emph{meta-generalization error} is defined as
\begin{equation} \label{eq:definition of meta-generalization error}
    \Delta_{\tau} \triangleq \mathbb{E}_{\mathcal{D}_{1:n}, \boldsymbol \theta }[\bar{F}_{\tau}(\mv \theta) - \hat{F}_{\mathcal{D}_{1:n}}(\mv \theta)],
\end{equation}
where the average is over the distribution of the global data set $\mathcal{D}_{1:n}$ and of the \textcolor{black}{pre-trained vector} $\mv \theta$ produced by the federated learning process based on the global data set $\mathcal{D}_{1:n}$. 
We write the conditioned distribution of the \textcolor{black}{pre-trained vector} given the global data set $\mathcal{D}_{1:n}$ as $\mathrm{P}_{\boldsymbol{\theta} | \mathcal{D}_{1:n}}$. The conditioned distribution $\mathrm{P}_{\boldsymbol{\theta} | \mathcal{D}_{1:n}}$ describe the overall operation of the protocol.

An upper bound on the meta-generalization error $\Delta_{\tau}$ in \eqref{eq:definition of meta-generalization error} would provide a measure of the discrepancy between the meta-training loss in \eqref{eq:definition of R_D}, which is optimized by the protocol, and the target meta-test loss.
In light of this, in the following subsections, we aim to analyze the generalization of the proposed Air-meta-pFL by deriving an upper bound of the error defined in \eqref{eq:definition of meta-generalization error}.

\subsection{Assumptions}
The generalization analysis requires the following additional assumptions.
\begin{assumption}[Sub-Gaussian Per-task Training Loss~\cite{jose2021information}] \label{assumption:sub-gaussian per-task loss}
    For all hyperparameters $\boldsymbol{\theta} \in \mathbb{R}^{d}$, the \emph{per-task training loss} function 
    \begin{equation}
        L(\mv \theta | \mathcal{D}) \triangleq \frac{1}{m^{(\rm{va})}}\sum_{Z\in \mathcal{D}^{(\rm{va})}} \ell \bigg(\boldsymbol{\theta} - \frac{\alpha }{m^{(\rm tr)}} \sum_{Z \in \mathcal{D}^{(\rm tr)}} \nabla \ell(\mv \theta; Z); Z \bigg)
    \end{equation}
    is $\sigma^2$-sub-Gaussian, where randomness is due to the dependence on the local data set $\mathcal{D}\sim \mu^m$.
\end{assumption}

Assumption \ref{assumption:sub-gaussian per-task loss} is automatically satisfied with $\sigma^2 = (b-a)^2/4$ if the loss function $\ell(\cdot ;\cdot)$ satisfies the inequality $a \le \ell(\cdot; \cdot) \le b$. 
\textcolor{black}{In practice, one can always satisfy the sub-Gaussian assumption by clipping the original loss or by applying a composition with a squashing function such as sigmoid function \cite{rivasplata2012subgaussian}.}

\begin{assumption}[Independent Mini-Batches] \label{assumption:sampling strategy}
The mini-batch sampling strategy for calculating the SGD in \eqref{eq:local updates} is such that the selected mini-batch is independent of the hyperparameter $\mv \theta_{i}^{(t, q)}$, and of mini-batches of previous rounds for all $t \in \{0,1,...,T-1\}$, $q\in \{0,1,...,Q-1\}$ and all devices $i\in [n]$.
\end{assumption}

\begin{assumption}[Bounded Sparsified Updates] \label{assumption:lower bound of gradient}
    For all $i\in [n]$ and $t \in \{0,1,...,T-1\}$, there exists a constant $\epsilon_g > 0$ such that the expected square norm of the sparsified update \eqref{eq:Comp-k} satisfies the inequality $\mathbb{E}\| \mv g_i^{(t)} \|^2 \ge \eta^2 \epsilon_g$, where $\eta$ is the learning rate in \eqref{eq:local updates}.
\end{assumption}

\subsection{Generalization Analysis} \label{subsec:generalization analysis}
To start, we review the following lemma derived in \cite{jose2021information}, which bounds the meta-generalization error in \eqref{eq:definition of meta-generalization error} for any protocol described by a conditioned distribution $\mathrm{P}_{\boldsymbol{\theta}|\mathcal{D}_{1:n}}$ via the mutual information $I(\mv \theta ; \mathcal{D}_{1:n})$ between the \textcolor{black}{pre-trained vector} $\mv \theta$ and the training data set $\mathcal{D}_{1:n}$.
{\color{black}
Note that conventional techniques to analyze generalization performance, such as VC dimension or Rademacher complexity, yield bounds that are determined by the complexity of the model class rather than training algorithms. 
By comparison, the mutual information is algorithm-dependent \cite{chen2021generalization}, yielding stronger design insights.
}

\begin{lemma}[Theorem 1, \cite{jose2021information}] \label{lemma:MI bound of meta learning}
    Under Assumption~\ref{assumption:sub-gaussian per-task loss}, for any protocol, characterized by a conditional distribution ${\mathrm{P}}_{\boldsymbol{\theta}\mid \mathcal{D}_{1:n}}$, the meta-generalization error $\Delta_{\tau}$ in \eqref{eq:definition of meta-generalization error} satisfies the inequality 
    \begin{equation} \label{eq:MI bound of meta learning}
        |\Delta_{\tau}| \le \sqrt{\frac{2\sigma^2}{n} I(\mv \theta; \mathcal{D}_{1:n}) }.
    \end{equation}
\end{lemma}

Lemma~\ref{lemma:MI bound of meta learning} suggests that a smaller correlation between the pre-trained vector $\mv \theta$ and training data sets $\mathcal{D}_{1:n}$, and thus smaller mutual information $I(\mv \theta; \mathcal{D}_{1:n})$, improve the generalization performance.
This is because a smaller correlation entails a more limited sensitivity to the specific realization of the training data.

In the sequel, we derive such a specific form, revealing the unique role played by the communication parameters in generalization that distinguishes itself from in convergence.

\begin{theorem}[Generalization of Air-meta-pFL] 
\label{theorem:generalization Air-meta-pFL}
    Under Assumptions~\ref{assumption:Bounded Variance and Norm}, \ref{assumption:gradient-Lipschitz}, \ref{assumption:Estimation Error}, \ref{assumption:sub-gaussian per-task loss}, \ref{assumption:sampling strategy} and \ref{assumption:lower bound of gradient}, the meta-generalization error of Air-meta-pFL satisfies the following inequality  
    \begin{equation} \label{eq:generalization bound with comp}
        |\Delta_{\tau}|
        \le \sqrt{ \frac{d \sigma^2}{n}\sum_{t=0}^{T-1}\log\left(1 + \frac{M P_{\max} rn C_g \sum_{i\in\mathcal{I}^{(t)}}|h_i^{(t)}|^2}{d v^{(t)} \epsilon_g} \right) },
    \end{equation}
    where
    $
        C_g=4Q^2 G^2 (\Lambda + 1)\left( (1+\alpha L_{G})^2 + \frac{\alpha^2 \sigma_H^2}{m_B} \right),
    $
    $P_{\max}=\max_{i\in[n]}P_i$, and $\Lambda = \frac{(1-\lambda)(1+1/c)}{1-(1-\lambda)(1+c)}$ with $\lambda=k/d$ and $0<c<\frac{\lambda}{1-\lambda}$.
\end{theorem}
\begin{IEEEproof}
    See Appendix~\ref{appendix:proof of generalization Air-meta-pFL}.
\end{IEEEproof}

{\color{black}
\begin{remark}
    To bound the mutual information $I(\mv \theta; \mathcal{D}_{1:n})$, we need to bound the conditional entropy $h(\mv \theta^{(t)} \mid \mv \theta^{(t-1)})$ by data processing inequality (c.f. \eqref{eq:MI intermediate bound on entropy}, Appendix~\ref{appendix:proof of generalization Air-meta-pFL}). Conditioned on $\mv \theta^{(t-1)}=\mv \vartheta^{(t-1)}$, $h(\mv \theta^{(t)} \mid \mv \theta^{(t-1)}=\mv \vartheta^{(t-1)})$ is bounded by the entropy of a Gaussian random variable with the same covariance, i.e., $\mv \theta^{(t)}-\mv \vartheta^{(t-1)}$. 
    As a result, the key step remains to bound the norm square $\mb{E}\|\mv \theta^{(t)} - \mv \vartheta^{(t-1)}\|^2$. However, this is technically challenging as it involves bounding algorithm-specific parameters including the square norm of the mini-batch gradient $\mathbb{E}\|\hat\nabla F_i(\mv\theta_i^{(t,q)})\|^2$, the memory vector $\mathbb{E}\|\mv m_i^{(t)}\|^2$ and the power scaling factor $ \rho^{(t)}$ (c.f. (67), Appendix~\ref{appendix:proof of generalization Air-meta-pFL}), none of which is covered by general results provided in [42] and [43] and thus requires exclusive analysis via Lemma \ref{lemma:bound of SGD F}, \ref{lemma:memory bound} and \ref{lemma:power scaling factor bound}. 
\end{remark}
}

The bound in \eqref{eq:generalization bound with comp}, \textcolor{black}{while generally not tight,} provides insights into the impact of the number of active devices, of SGD, of the available communication resources, and of data heterogeneity on generalization.
On one hand, like the convergence bound \eqref{eq:convergence bound}, the generalization bound in \eqref{eq:generalization bound with comp} increases with the variance $\sigma_F^2$ of the outer stochastic gradient, as well as with the level of data heterogeneity via the parameter $\gamma_F^2$. 
However, on the other hand, the impact of the communication parameters on generalization is significantly different from that revealed on convergence.

For example, increasing the number of channel uses $M$ compromises generalization, while improving on convergence performance by reducing estimation error (c.f.~\eqref{eq:convergence bound}). 
Meanwhile, setting $P_{\min}=P_{\max}=P$ for simplicity, the generalization bound in \eqref{eq:generalization bound with comp} demonstrates that a small SNR $P|h_i^{(t)}|^2/v^{(t)}$ decreases the generalization bound at the cost of deteriorating the convergence via estimation error.
Intuitively, this result stems from the fact that an increased disturbance on the communication channel decreases the correlation between training data and \textcolor{black}{pre-trained vector} $\mv \theta$, possibly reducing overfitting.

\section{Numerical Results} \label{sec:Numerical Results}
\begin{table}[t]
    \caption{Experimental parameters}
    \centering
    \begin{tabular}{llll}
        \hline
        \textbf{Parameter}             & \textbf{Value}                                                                                               & \textbf{Parameter}             & \textbf{Value}                                                          \\ \hline
        Mini-batch size                & $m_B=32$                                                                                                     & $N$-way                        & $N=5$                                                                   \\ \hline
        Local SGD                      & \begin{tabular}[c]{@{}l@{}}$Q=5$ for convergence analysis\\ $Q=1$ for generalization analysis\end{tabular}   & $K$-shot                       & $K=8$                                                                   \\ \hline
        Number of devices for training & $n=9$                                                                                                        & Compression rate               & $M/d=0.4$                                                               \\ \hline
        Number of devices for test     & $n_{\text{test}}=3$                                                                                          & Sparsification rate            & $k/d=0.04$                                                              \\ \hline
        Fraction of active device      & \begin{tabular}[c]{@{}l@{}}$r=0.3$ for convergence analysis\\ $r=1$ for generalization analysis\end{tabular} &  Linear compression matrix $\mv A$ &  $M$ out of $d$ rows DFT matrix  \\ \hline
        Number of global rounds        & $T=200$                                                                                                      & Channel fading                 & Rayleigh fading                                                         \\ \hline
        Meta-learning rate             & $\eta^{(t)}=\eta=0.4$                                                                                        & Estimator $\mathcal{E}(\cdot)$ & OAMP with $20$ iterations~\cite{ma2017orthogonal} \\ \hline
        Inner learning rate            & $\alpha^{(t)}=\alpha=0.4$                                                                                    & Max. \# Monte Carlo trials     & 100                                                                     \\ \hline
        \end{tabular}   \label{table:experimental parameters}
\end{table}

In this section, numerical experiments are conducted to validate the performance of the proposed Air-meta-pFL protocol, as well as the convergence and generalization analysis presented in the previous sections.

\subsection{Experimental Settings} \label{subsec:Experimental Settings}
We adopt the Omniglot data set, which contains $1643$ characters, considered as classes, with each class having $p=20$ instances drawn by different persons, amounting to $32860$ instances in total~\cite{finn2017model,chen2021generalization,lake2011one}. 
We employ $n=9$ devices for pFL training and $n_{\text{test}}=3$ additional devices for evaluation of the generalization performance. All devices are assigned non-$i.i.d.$ local data sets. 

\begin{figure}[t]
    \centering
    \includegraphics[width=1\linewidth]{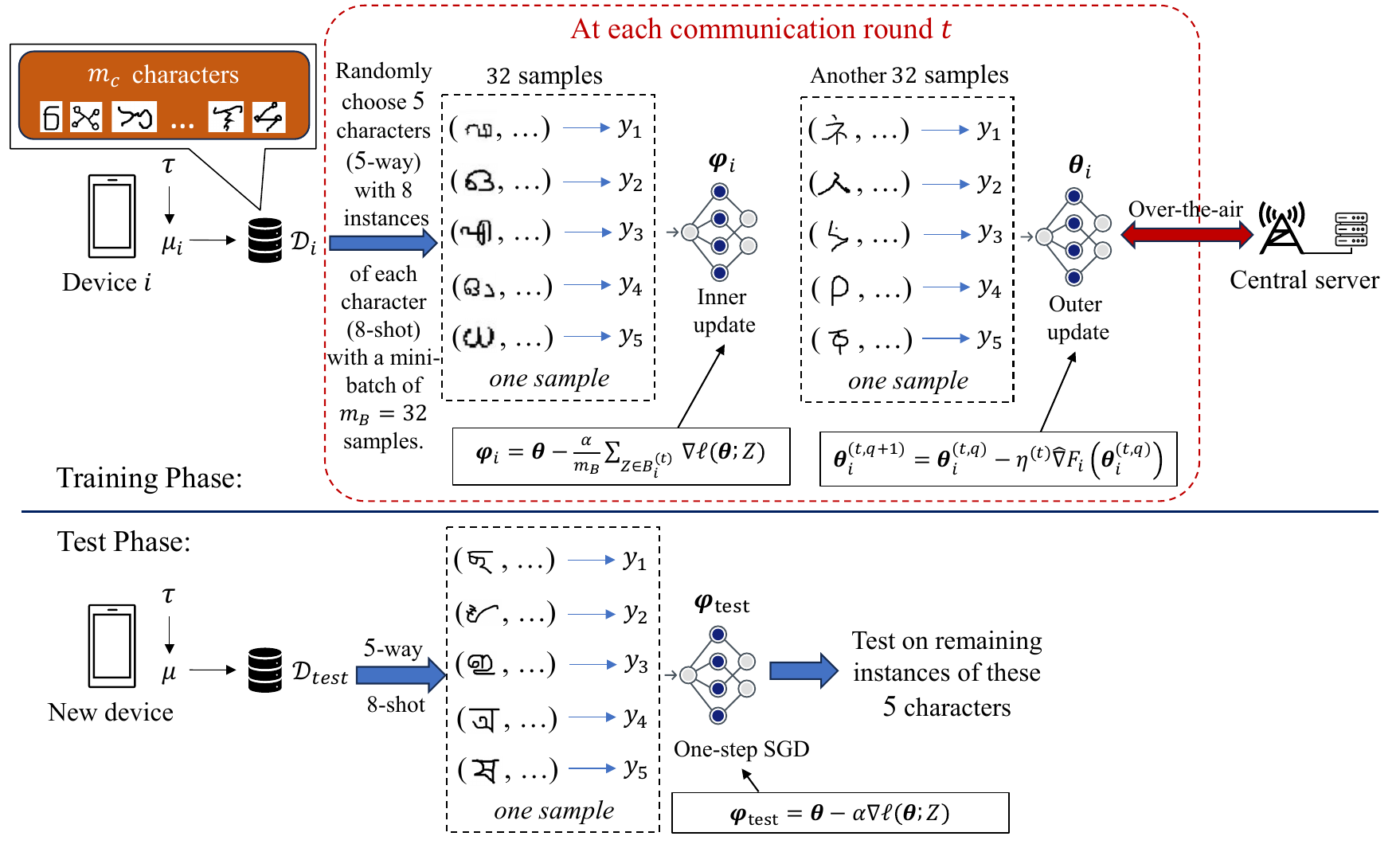}
  \caption{ \textcolor{black}{Illustration of the experimental setting for the Omniglot data set ($N=5, K=8, m_B=32, m_c=136$ or $10$).} }
    \label{fig:experiment_illustration}
\end{figure}

\textcolor{black}{The experimental setting is illustrated in Fig.~\ref{fig:experiment_illustration}.}
Specifically, our experiments adopt the $N$-way $K$-shot classification protocol as follows~\cite{finn2017model}. 
{\color{black}
Note that first-order MAML achieves comparable test accuracy as MAML for $N$-way $K$-shot tasks~\cite[Table 1]{finn2017model} (see Fig.~\ref{fig:acc_vs_t}), which provides evidence that the proposed method with first-order approximation is potentially applicable to large models.
}
The task of each training device $i \in [n]$ is to distinguish among $N$ different classes, selected out of a set of $0\le m_c \le 1643$ classes specific to each device, given $K$ instances of each of the $N$ classes. 
Performance is then assessed by the test accuracy for classification of instances on newly $N$ selected classes within the device-specific $m_c$ classes. 
For each training device $i \in [n]$, $m_c$ different characters (classes) are drawn uniformly without replacement from Omniglot. 
The data set $\mathcal{D}_i$ includes $p=20$ instances for each of the $m_c$ classes. 
The local data sets for the $n_{\text{test}}$ test devices are generated following the same procedure.
Accordingly, the level of data heterogeneity across devices decreases with $m_c$.
In our experiment, we set $N=5$ and $K=8$. At each communication round $t$, each active device $i \in \mathcal{I}^{(t)}$ randomly chooses a mini-batch $m_B=32$ $N$-way $K$-shot tasks from data set $\mathcal{D}_i$ for inner update and another $m_B=32$ tasks for outer update. 
The tasks are obtained by selecting $N$ classes uniformly at random from the $m_c$ available classes.
The meta-test loss $\bar{F}_{\tau}(\mv \theta)$ and test accuracy are averaged over $1000$ different $N$-way $K$-shot tasks of each test device, and over $n_{\text{test}}$ test devices. 

We employed a CNN network architecture comprising several modules. The initial three modules share identical structures, each featuring a $3 \times 3$ 2D convolution layer with 64 filters and a stride of 2, followed by a ReLU activation layer and batch normalization (BN). The fourth module consists of a $2 \times 2$ 2D convolution layer with 64 filters and a stride of 1, along with a Relu activation layer and BN.  This feature map with shape $64 \times 1 \times 1$ is then fed into a fully connected layer with $5$ nodes. This CNN model results in $d=91781$ parameters. 

For the communication setup, we consider the block-flat Rayleigh fading channel, i.e., $h_i^{(t)}\sim \mathcal{CN}(0,1)$ in \eqref{eq:rxsignal}, and control the received SNR to be approximately equal to $19$\,dB. The number of available channel uses for each device is set as $M=0.4 d$. 
\textcolor{black}{The linear compression matrix $\mv A^{(t)}$ is generated by randomly selecting $M$ out of $d$ rows of a $d$-by-$d$ discrete Fourier transform (DFT) matrix.}
Unless otherwise noted, the experimental parameters of the Air-meta-pFL are summarized in Table~\ref{table:experimental parameters}. 

\subsection{Convergence Analysis} \label{subsec:experiment convergence}
\begin{figure}[t]
    \centering
    \includegraphics[width=3.2in]{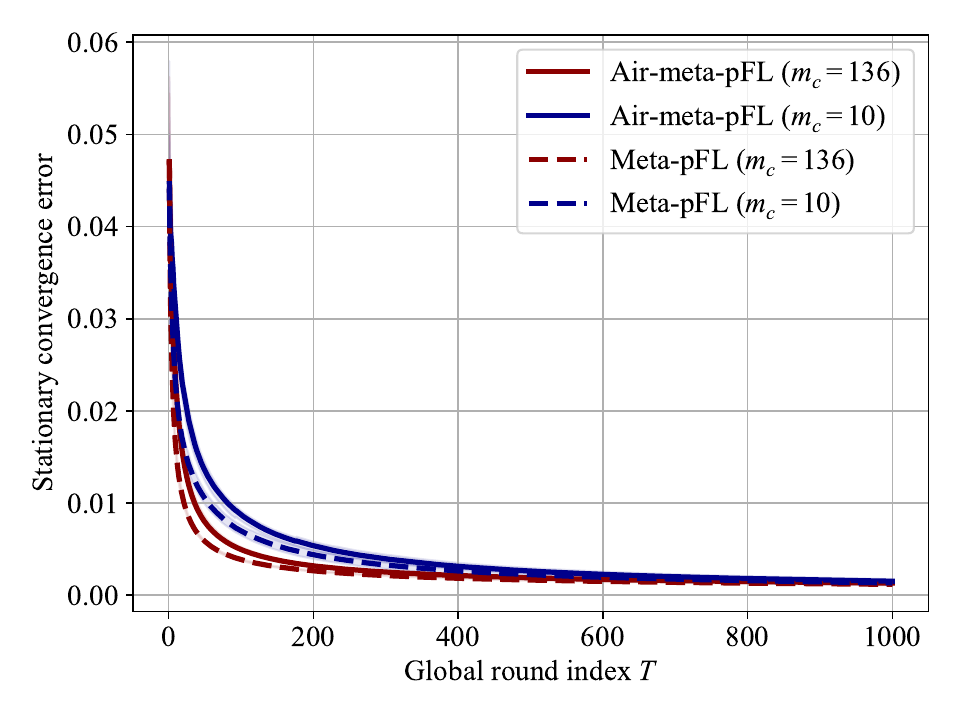}
    \caption{Stationary convergence error, i.e., square norm of the gradient of the meta-training loss, versus global round $t$ for meta-pFL, which assumes ideal communication, and for Air-meta-pFL.  The shaded error bars correspond to intervals covering 95\% of the realized values, obtained from the $10$ Monte Carlo trials.}
    \label{fig:loss_vs_t}
\end{figure}

We first study the convergence to stationary solution of the meta-training loss by evaluating the average square norm gradients of the meta-training loss, $\Myfrac{1}{T}\sum_{t=0}^{T-1}\mathbb{E}\|\nabla F(\mv \theta^{(t)}) \|^2$, versus the communication round $T$ with two different values of $m_c$ in Fig.~\ref{fig:loss_vs_t}. 
We refer to this quantity as the stationary convergence error. 
It is observed that the stationary convergence error of Air-meta-pFL decreases to a value similar to PerFedAvg, which assumes ideal communications. 
Moreover, as predicted by Theorem~\ref{theorem:convergence}, a larger heterogeneity, i.e., smaller $m_c$, leads to a slower convergence.

\begin{figure}[t]
    \centering
    \begin{minipage}[t]{0.48\textwidth} 
        \centering
        \includegraphics[width=3.2in]{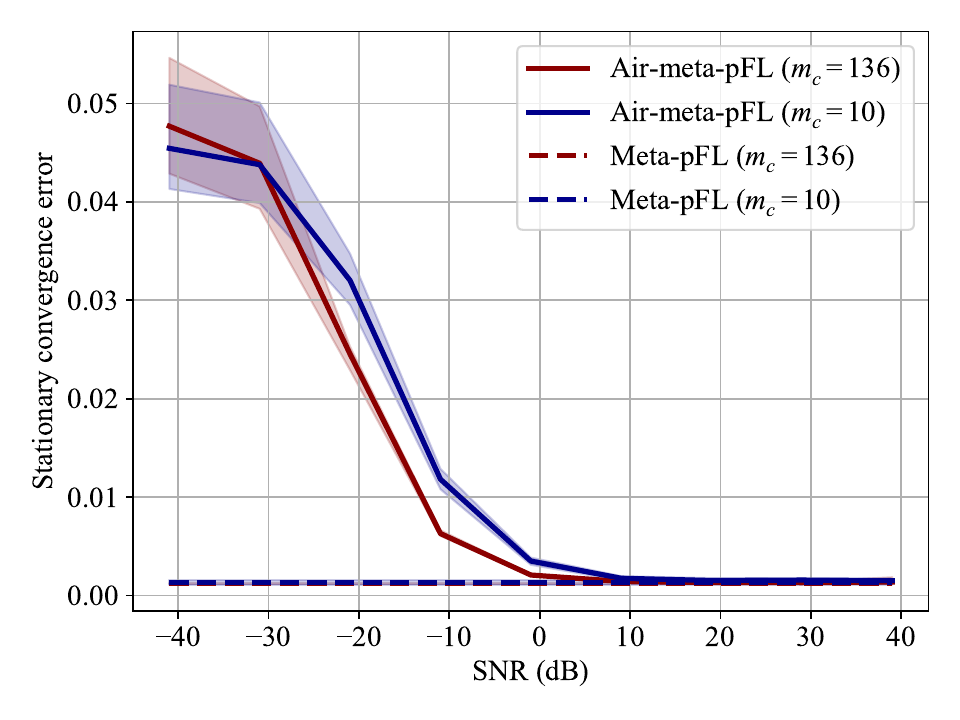}
        \caption{Stationary convergence error versus received SNR for Air-meta-pFL and meta-pFL. }
        \label{fig:grad_vs_snr}
    \end{minipage}
    \hfill
    \begin{minipage}[t]{0.48\textwidth} 
        \centering
        \includegraphics[width=3.35in]{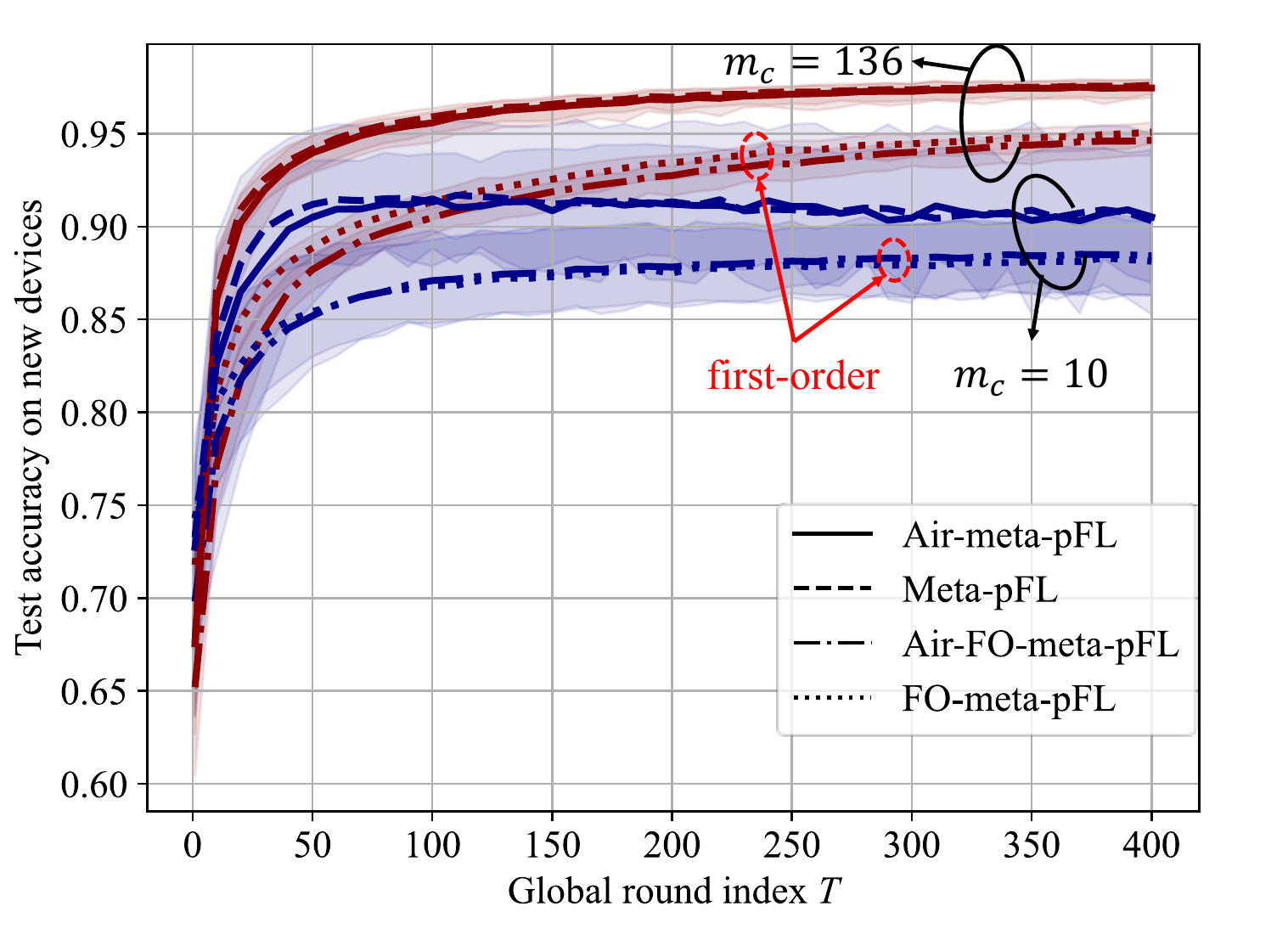}
        \caption{{\color{black}Test accuracy versus global round $t$ for Air-meta-pFL, meta-pFL and their first-order variants, Air-FO-meta-pFL and FO-meta-pFL.}}
        \label{fig:acc_vs_t}
    \end{minipage}
\end{figure}

We further evaluate the stationary convergence error as a function of the received SNR in Fig.~\ref{fig:grad_vs_snr}. In line with Theorem~\ref{theorem:convergence}, the error decreases with the SNR, yielding the same performance as the idealized meta-pFL scheme.


{\color{black}
Finally, we compare the test accuracy on the test devices for Air-meta-pFL, meta-pFL, and their first-order variants, Air-FO-meta-pFL and FO-meta-pFL, versus the communication round $T$ in Fig.~\ref{fig:acc_vs_t}. 
}
Air-meta-pFL is observed to achieve performance comparable to meta-pFL despite the limitations imposed by wireless communication. 
As suggested by the generalization bound and the convergence bound, Fig.~\ref{fig:acc_vs_t} shows a large heterogeneity level, i.e., a smaller $m_c$, which deteriorates the learning performance. 
{\color{black}
Furthermore, Fig.~\ref{fig:acc_vs_t} reveals that the test accuracy degradation induced by the first-order approximation is less than $3$\% compared to the original algorithms.
}


\subsection{Generalization Analysis} \label{subsec:experiment generalization}
Next, we conduct experiments to validate the insights obtained by the generalization analysis in Theorem~\ref{theorem:generalization Air-meta-pFL}. 
To this end, the meta-generalization error $\Delta_{\tau}$ in \eqref{eq:definition of meta-generalization error} is calculated as the difference between the expected meta-test loss $\mathbb{E}_{\boldsymbol{\theta}}[\bar{F}_{\tau}(\mv \theta)]$ and the expected meta-training loss $\mathbb{E}_{\mathcal{D}_{1:n}, \boldsymbol{\theta}}[\hat{F}_{\mathcal{D}_{1:n}}(\mv \theta)]$.
The meta-training loss $\hat F_{\mathcal{D}_{1:n}}(\mv\theta)$ and the meta-test loss $\bar{F}_{\tau}(\mv \theta)$ are both averaged over the shared parameter $\mv \theta$ via Monte Carlo trials.

\begin{figure}[t]
    \centering
    \begin{minipage}[t]{0.48\textwidth} 
        \centering
        \includegraphics[width=3.2in]{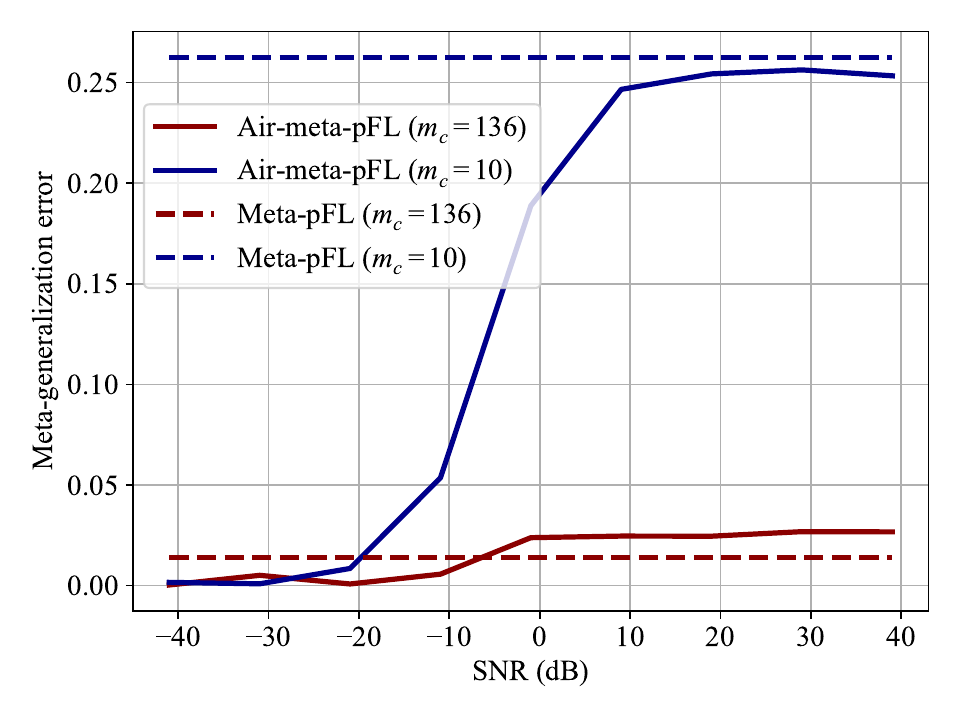}
        \caption{Meta-generalization error $|\Delta_\tau|$ versus the received SNR for Air-meta-pFL and meta-pFL.}
        \label{fig:gen_vs_power_omniglot}
    \end{minipage}
    \hfill
    \begin{minipage}[t]{0.48\textwidth} 
            \centering
            \includegraphics[width=3.2in]{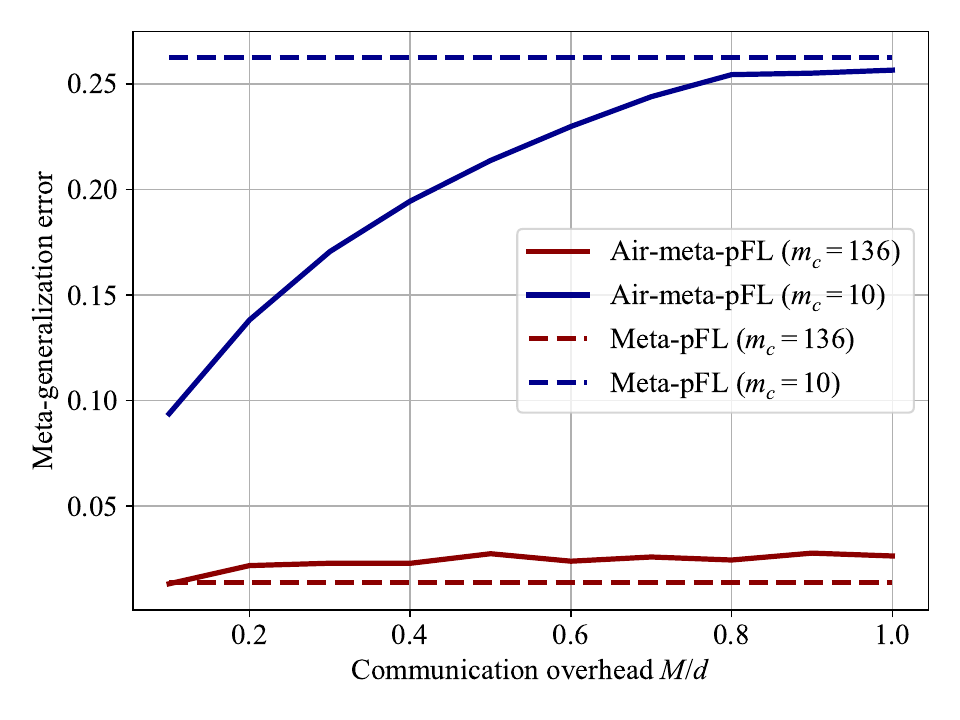}
            \caption{Meta-generalization error $|\Delta_\tau|$ versus communication overhead $M/d$ with compression rate equal to $k/d=0.06$ for Air-meta-pFL and meta-pFL.}
            \label{fig:gen_vs_M/d_omniglot}
    \end{minipage}
\end{figure}



We focus on the impact of communication impairments on the generalization error $\Delta_{\tau}$ for different levels of heterogeneity.
Fig.~\ref{fig:gen_vs_power_omniglot} shows the absolute value of meta-generalization error $|\Delta_{\tau}|$ versus the received SNR, with the ratio between the number of preserved parameters via top-$k$ and the number of parameters given by $k/d=0.06$ and the ratio between the number of channel uses and the number of parameters given by $M/d=0.8$. 
This result shows that, in line with Theorem~\ref{theorem:generalization Air-meta-pFL}, increasing the received SNR entails a larger meta-generalization error $\Delta_{\tau}$, which also increases with the level of heterogeneity. 
Thus, communication noise will potentially mitigate overfitting.
In fact, for a large $m_c$, Air-meta-pFL may even outperform meta-pFL in terms of generalization, owing to the beneficial effects of channel noise.

Fig.~\ref{fig:gen_vs_M/d_omniglot} shows the absolute value of meta-generalization error, $|\Delta_{\tau}|$, versus the communication overhead rate $M/d$ with $k/d=0.06$.     
Confirming the prediction of Theorem~\ref{theorem:generalization Air-meta-pFL}, increasing $M/d$ leads to smaller estimation errors, allowing Air-meta-pFL to approach, and potentially surpass, the performance of meta-pFL.

\subsection{Convergence-Generalization Trade-Off} \label{subsec:experiment trade-off}
\begin{figure*}[t]
    \centering
    \subfigure[]{
        \centering
        \includegraphics[width=3in]{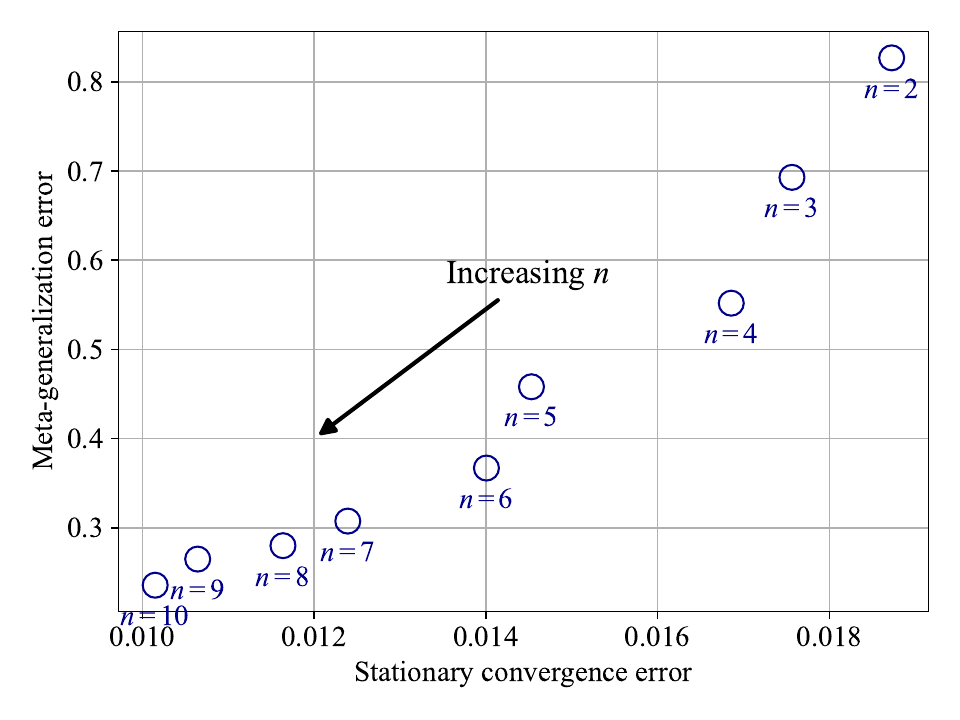}
        \label{fig:trade_off_vs_n_omniglot}
    }\hfill
    \subfigure[]{
        \centering
        \includegraphics[width=3in]{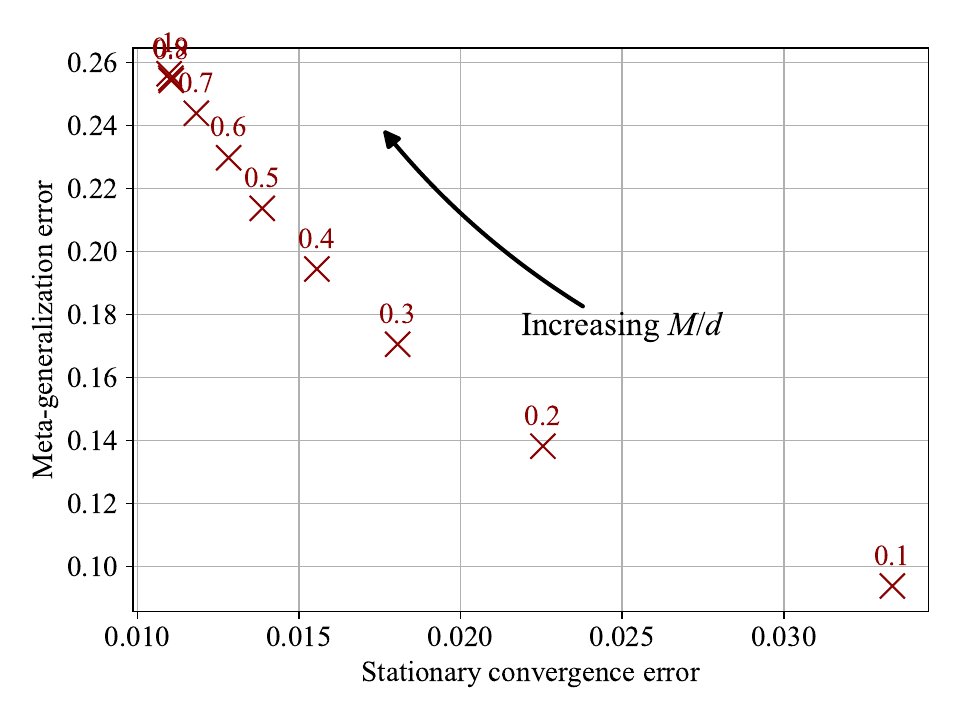}
        \label{fig:trade_off_vs_M/d_omniglot}
    }
    \caption{Generalization-convergence trade-off by varying: (a) the number of training devices $n$; and (b) the communication overhead $M/d$ with $m_c=10$ averaged over 10 Monte Carlo trials.}
    \label{fig:trade_off}
\end{figure*}

In this subsection, we finally conduct experiments to elucidate the convergence-generalization trade-off. 
To this end, Fig.~\ref{fig:trade_off} shows the meta-generalization error, $|\Delta_{\tau}|$, versus the stationary convergence error, $\Myfrac{1}{T}\sum_{t=0}^{T-1}\mathbb{E}\|\nabla F(\mv \theta^{(t)}) \|^2$, where we set $m_c=10$.
In Fig.~\ref{fig:trade_off_vs_n_omniglot}, the trade-off is traced by varying the number of training devices $n$, while in Fig.~\ref{fig:trade_off_vs_M/d_omniglot}, we vary the communication overhead $M/d$.
{\color{black}
From Fig.~\ref{fig:trade_off_vs_n_omniglot}, as indicated by Theorem~\ref{theorem:convergence} and Theorem~\ref{theorem:generalization Air-meta-pFL}, increasing the number of training devices $n$ benefits both generalization and convergence.
As shown in Table~\ref{table:acc_vs_n_response}, the test accuracy is seen to increase with the number of clients $n$, demonstrating the scalability and efficiency of the proposed Air-meta-pFL.
}
In contrast, from Fig.~\ref{fig:trade_off_vs_M/d_omniglot}, as also reflected by the theorems, the meta-generalization error increases, while the stationary convergence error decreases with $M/d$. 
This highlights the different roles played by the communication parameters on convergence and generalization. 
In this regard, communication impairments have a regularizing effect on generalization, while negatively affecting convergence.

\begin{table}[ht]
    \caption{\textcolor{black}{Test accuracy versus different numbers of devices $n$ for Omniglot}}
    \centering
    \begin{tabular}{llllllllll}
        \hline
        $n$ & 2 & 3 & 4 & 5 & 6 & 7 & 8 & 9 & 10 \\ \hline
 Test accuracy & 0.855 & 0.882 & 0.887 & 0.901 & 0.906 & 0.919 & 0.919 & 0.922 & 0.923
        \\ \hline
        \end{tabular}   \label{table:acc_vs_n_response}
\end{table}

\section{Conclusion and Discussion} \label{sec:conclusion}

In this paper, we studied a wireless implementation of meta-pFL, a formalized framework of customizing pre-trained models to new agents and tasks using distributed data sets in a federated way. Aiming for efficient use of the shared wireless resources, we introduced a wireless protocol that leverages over-the-air computing,  model sparsification, and linear compression, along with an error-compensation mechanism. The main result of this paper is the derivation of bounds on convergence and generalization error, with generalization referring to the performance upon fine-tuning on new users and tasks. The analysis offers insights into the trade-off between convergence and generalization, as wireless impairments may compromise convergence, while potentially enhancing generalization. 
{\color{black}
For example, a larger transmission power leads to convergence speedups, while potentially degrading generalization performance.
}

Future work will investigate these trade-offs in scenarios involving more complicated training algorithms such as fully decentralized implementations, asynchronous updates, and fine-tuning of LLMs, as well as large action models (LAMs). For instance, recent advancements in orthogonal time frequency and space (OTFS) modulation potentially improve performance in dynamic environments like linear time-varying (LTV) channels \cite{rakesh2025over}. 
{\color{black}
Another interesting future direction is to optimize transmit power and the number of channel uses to improve on the convergence-generalization trade-offs. 
}

\appendix
\subsection{Important Lemmas}~\label{appendix:Important Lemmas}
In this subsection we provide several important lemmas for the proofs. 

\begin{lemma}[Young's inequality]~\label{lemma:Youngs inequality}
    For any $\mv a, \mv b \in \mathbb{R}^{d}$ and $\epsilon > 0$, we have 
    \begin{equation*}
        |\left<\mv a,\mv b\right>| \le \frac{\epsilon}{2}\| \mv a \|^2 + \frac{1}{2\epsilon}\| \mv b\|^2.
    \end{equation*}
\end{lemma}

\begin{lemma}[Lemma 4.2 in~\cite{fallah2020personalized}]~\label{lemma:L-smooth of F}
    Suppose that Assumption~\ref{assumption:Bounded Variance and Norm} and \ref{assumption:gradient-Lipschitz} are satisfied and $\alpha \in (0, 1/L_{G}]$, the local loss $F_i(\mv \theta)\triangleq f_i(\mv \theta - \alpha \nabla f_i(\mv \theta))$ is smooth with parameter $L_F = 4L+\alpha L_H G$.
\end{lemma}

\begin{lemma}[Lemma 4.3 in~\cite{fallah2020personalized}]~\label{lemma:Bounded variance of F}
    The estimate $\hat \nabla F_i(\mv \theta)$ (c.f.~\eqref{eq:F_SGD}) is computed using independent batches with the same size  $m_B$. Suppose that Assumption~\ref{assumption:Bounded Variance and Norm} and \ref{assumption:gradient-Lipschitz} satisfied, then for any $\alpha \in (0, 1/L_{G}]$, $i\in[n]$ and $\mv w \in \mathbb{R}^{d}$, we have 
    \begin{equation*}
        \left\|\mathbb{E}\left[\hat{\nabla} F_i(\mv w)-\nabla F_i(\mv w)\right]\right\|^2 \leq \frac{4 \alpha^2 L^2_{G} \sigma^2_G}{m_B}, 
    \end{equation*}
    and 
    \begin{align*}
       \mathbb{E}\left[\left\|\hat{\nabla} F_i(\mv w)-\nabla F_i(\mv w)\right\|^2\right] \leq \sigma_F^2\triangleq 12\sigma_G^2\left(\frac{1}{m_B}+\frac{(\alpha L_{G})^2}{m_B}\right)\left(1+\sigma_H^2 \frac{\alpha^2}{4 m_B}\right)+12 G^2 \sigma_H^2 \frac{\alpha^2}{4 m_B},
    \end{align*}
    where the expectation is over the randomness of SGD.
\end{lemma}

\begin{lemma}[Lemma 4.4 in~\cite{fallah2020personalized}]~\label{lemma:Bounded Heterogeneity of F}
    Suppose that Assumptions~\ref{assumption:Bounded Variance and Norm}-\ref{assumption:Bounded Data Heterogeneity} are satisfied, for any $\alpha \in (0,1/L_{G}]$ and $\mv w \in \mathbb{R}^d$, we have 
    \begin{equation*}
        \left\|\nabla F_i(\mv w)-\nabla F(\mv w)\right\|^2 \leq \gamma_F^2:=3 G^2 \alpha^2 \gamma_H^2+192 \gamma_G^2.
    \end{equation*}
\end{lemma}

\begin{lemma}~\label{lemma:bound of SGD F}
    For all $i \in [n]$, $t\in\{0,1,\ldots,T-1\}$ and $q\in\{0,1,\ldots,Q-1\}$, with independent mini-batches $\mathcal{B}_{i}^{(t, q)}, \mathcal{B}_{i}^{\prime(t, q)}$, and $\mathcal{B}_{i}^{\prime\prime(t, q)}\subseteq\mathcal{D}_{i}$ of the same size $m_B$, the expected square norm of $\hat{\nabla} F_{i}(\mv \theta_{i}^{(t, q)})$ is bounded by
    \begin{equation*}
        \mathbb{E}\| \hat{\nabla} F_i(\mv \theta_i^{(t,q)}) \|^2 \le 2\left((1+\alpha L_{G})^2 + \frac{\alpha^2 \sigma_H^2}{m_B}\right)G^2.
    \end{equation*}
\end{lemma}
\begin{IEEEproof}
    For any $i\in[n]$ and $\mv \theta \in \mathbb{R}^d$, we have
    \begin{align} \nonumber
            \| \hat{\nabla} F_i(\mv \theta_i^{(t,q)}) \| 
            & = \bigg\|(\mv I_d - \alpha \nabla^2 f_i(\mv \theta_i^{(t,q)}) + \mv e_i) \hat{\nabla} f_i\left(\mv \theta_{i}^{(t, q)} - \alpha \nabla f_{i}(\mv \theta_{i}^{(t, q)}, \mathcal{B}_{i}^{(t, q)}), \mathcal{B}_{i}^{\prime(t, q)}\right)\bigg\|
            \nonumber \\ 
            & \le \|\mv I_d - \alpha \nabla^2 f_i(\mv \theta_i^{(t,q)}) + \mv e_i \|_2 \left\|\hat{\nabla} f_i\left(\mv \theta_{i}^{(t, q)} - \alpha \nabla f_{i}(\mv \theta_{i}^{(t, q)}, \mathcal{B}_{i}^{(t, q)}), \mathcal{B}_{i}^{\prime(t, q)}\right)\right\|, 
    \end{align}
    where $\mv e_i = \alpha(\nabla^2 f_i(\mv \theta_i^{(t,q)}) - \hat{\nabla}^2 f_i(\mv \theta_i^{(t,q)}, \mathcal{B}_{i}^{\prime\prime(t,q)}))$. By Assumption~\ref{assumption:Bounded Variance and Norm}, $\mathbb{E}\|\mv e_i \|_2^2 \le \Myfrac{\alpha^2 \sigma_H^2}{m_B}$. Then we have, 
    \begin{align}
            & \mathbb{E}\| \hat{\nabla} F_i(\mv \theta_i^{(t,q)}) \|^2 \\ 
            & \overset{(a)}{\le} \mathbb{E}\bigg[ \left(2\|\mv I_d - \alpha \nabla^2 f_i(\mv \theta_i^{(t,q)})\|_2^2 + 2\|\mv e_i \|_2^2 \right)  \left\|\hat{\nabla} f_i\left(\mv \theta_{i}^{(t, q)} - \alpha \nabla f_{i}(\mv \theta_{i}^{(t, q)}, \mathcal{B}_{i}^{(t, q)}), \mathcal{B}_{i}^{\prime(t, q)}\right)\right\|^2 \bigg] \nonumber\\ 
            & \overset{(b)}{=} \left(2\|\mv I_d - \alpha \nabla^2 f_i(\mv \theta_i^{(t,q)})\|^2 + 2\mathbb{E}\|\mv e_i \|^2 \right) \mathbb{E}\left\|\hat{\nabla} f_i\left(\mv \theta_{i}^{(t, q)} - \alpha \nabla f_{i}(\mv \theta_{i}^{(t, q)}, \mathcal{B}_{i}^{(t, q)}), \mathcal{B}_{i}^{\prime(t, q)}\right)\right\|^2 \nonumber \\ 
            & \overset{(c)}{\le} 2\left((1+\alpha L_{G})^2 + \frac{\alpha^2 \sigma_H^2}{m_B}\right)G^2,
    \end{align}
    where (a) is followed by Jensen's inequality, (b) is due to the independence between $\mathcal{B}_{i}^{(t, q)}, \mathcal{B}_{i}^{\prime(t, q)}$, and $\mathcal{B}_{i}^{\prime\prime(t, q)}$, (c) is follwed by Assumption~\ref{assumption:Bounded Variance and Norm} and \ref{assumption:gradient-Lipschitz}, and $\|\mv I_d - \alpha \nabla^2 f_i(\mv \theta_i^{(t,q)})\|_2 \le \|\mv I_d\|_2 + \|\alpha \nabla^2 f_i(\mv \theta_i^{(t,q)})\|_2 \le 1+\alpha L_{G}$.
\end{IEEEproof}

\begin{lemma}~\label{lemma:difference between local and global theta}
    For $\eta \le \frac{1}{10 Q L_F}$, we have 
    \begin{equation*}
        \mathbb{E}\| \mv \theta_{i}^{(t,q)} - \mv \theta^{(t)} \|^2 \le 40 Q^2 \eta^2 (\sigma_F^2 + \gamma_F^2) + 40Q^2 \eta^2 \mathbb{E}\| \nabla F(\mv \theta^{(t)})\|^2.
    \end{equation*}
\end{lemma}
\begin{IEEEproof}
    \begin{align}
        & \mathbb{E} \| \mv \theta_{i}^{(t,q)} - \mv \theta^{(t)} \|^2 \\ 
        & =  \mathbb{E}  \| \mv \theta_{i}^{(t,q-1)} - \mv \theta^{(t)} - \eta \hat{\nabla}F_i(\mv \theta_i^{(t,q-1)}) \|^2 \nonumber\\ 
        & \overset{(a)}{\le}  \left(1+\frac{1}{2Q - 1}\right)\mathbb{E} \| \mv \theta_{i}^{(t,q-1)} - \mv \theta^{(t)}\|^2  + 2Q \mathbb{E}\| \eta \hat{\nabla}F_i(\mv \theta_i^{(t,q-1)}) \|^2 \nonumber\\ 
        & =  \left(1+\frac{1}{2Q - 1}\right)\mathbb{E}\| \mv \theta_{i}^{(t,q-1)} - \mv \theta^{(t)}\|^2 + 2Q \eta^2 \mathbb{E} \big\|\hat{\nabla}F_i(\mv \theta_i^{(t,q-1)}) -  \nabla F_i(\mv \theta_i^{(t,q-1)}) + \nabla F_i(\mv \theta_i^{(t,q-1)}) \nonumber\\ 
        & \quad \quad \quad  + \nabla F_i(\mv \theta^{(t)}) - \nabla F_i(\mv \theta^{(t)}) + \nabla F(\mv \theta^{(t)}) - \nabla F(\mv \theta^{(t)}) \big\|^2 \nonumber\\ 
        & \overset{(b)}{\le} \left(1+\frac{1}{2Q - 1}\right)\mathbb{E}\| \mv \theta_{i}^{(t,q-1)} - \mv \theta^{(t)}\|^2 + 8Q\eta^2 \mathbb{E}\| \hat{\nabla}F_i(\mv \theta_i^{(t,q-1)}) -  \nabla F_i(\mv \theta_i^{(t,q-1)}) \|^2 \nonumber\\ 
        & \quad \quad + 8Q \eta^2 \bigg( \mathbb{E}\|\nabla F_i(\mv \theta_i^{(t,q-1)}) - \nabla F_i(\mv \theta^{(t)}) \|^2 + \|\nabla F_i(\mv \theta^{(t)}) - \nabla F(\mv \theta^{(t)}) \|^2 + \|\nabla F(\mv \theta^{(t)}) \|^2 \bigg) \nonumber\\ 
        & \overset{(c)}{\le} \left(1+\frac{1}{2Q - 1}+8Q\eta^2L_F^2\right)\mathbb{E}\| \mv \theta_{i}^{(t,q-1)} - \mv \theta^{(t)}\|^2 + 8Q\eta^2\sigma_F^2 + 8Q\eta^2\gamma_F^2 + 8Q\eta^2\|\nabla F(\mv \theta^{(t)})\|^2, 
\end{align}
where (a) and (b) are followed by Lemma~\ref{lemma:Youngs inequality}, (c) is due to Lemma~\ref{lemma:L-smooth of F}, Lemma~\ref{lemma:Bounded variance of F} and Lemma~\ref{lemma:Bounded Heterogeneity of F}. Let $\eta \le \frac{1}{10QL_F}$, which implies $\frac{1}{2Q - 1}+\frac{8}{100Q} \le \frac{1}{Q-1}$ for any $Q>1$. Then, we have 
\begin{align}
        & \mathbb{E} \| \mv \theta_{i}^{(t,q)} - \mv \theta^{(t)} \|^2 \nonumber\\  
        & \le \left(1+\frac{1}{Q-1}\right)\mathbb{E}\| \mv \theta_{i}^{(t,q-1)} - \mv \theta^{(t)}\|^2  + 8Q\eta^2\sigma_F^2 + 8Q\eta^2\gamma_F^2 + 8Q\eta^2\|\nabla F(\mv \theta^{(t)})\|^2 \nonumber\\ 
        & \le \sum_{q=0}^{Q-1}\left(1+\frac{1}{Q-1}\right)^q \left(8Q\eta^2(\sigma_F^2+\gamma_F^2 \right) + 8Q\eta^2\|\nabla F(\mv \theta^{(t)})\|^2) \nonumber\\ 
        & \le (Q-1)\left[\left(1+\frac{1}{Q-1}\right)^Q-1\right] \left(8Q\eta^2(\sigma_F^2+\gamma_F^2) + 8Q\eta^2\|\nabla F(\mv \theta^{(t)})\|^2 \right) \nonumber \\ 
        & \le 40Q^2\eta^2(\sigma_F^2+\gamma_F^2)+40Q^2\eta^2\|\nabla F(\mv \theta^{(t)})\|^2,
\end{align}
where the last inequality is due to $(1+\frac{1}{Q-1})^Q \le 5$ for all $Q>1$.
When $Q=1$, the bound is trivially satisfied since $\mv \theta_{i}^{(t,Q-1)} = \mv \theta_{i}^{(t,0)} = \mv \theta^{(t)}$.
\end{IEEEproof}

 \begin{lemma}\label{lemma:memory bound}
    For all $i\in[n]$ and $t\in\{0,1,\ldots,T-1\}$, the expcted suare norm of the memory vector \(\mv m_i^{(t)}\) is bounded by
    \begin{equation*}
        \mathbb{E}\|\mv m_i^{(t)}\|^2 \le 2\eta^2\Lambda Q^2\left( (1+\alpha L_{G})^2 + \frac{\alpha^2 \sigma_H^2}{m_B} \right)G^2,
    \end{equation*}
    where $\Lambda = \tfrac{(1-\lambda)(1+\frac{1}{c})}{1-(1-\lambda)(1+c)}$ and $0<c<\tfrac{\lambda}{1-\lambda}$ with \(\lambda=\tfrac{k}{d}\).
 \end{lemma}
 \begin{IEEEproof}
    \begin{align}
        & \mathbb{E}\| \mv m_i^{(t+1)} \|^2 \nonumber \\ 
        & = \mathbb{E}\| \mv m_i^{(t)} + \Delta_i^{(t)} - \mv g_i^{(t)} \|^2 \nonumber\\ 
        & \le (1-\lambda) \mathbb{E}\| \mv m_i^{(t)} + \Delta_i^{(t)} \|^2 \nonumber\\
        & \overset{(a)}{\le} (1-\lambda)(1+c)  \mathbb{E}\| \mv m_i^{(t)}\|^2 + (1-\lambda)(1+\frac{1}{c})\mathbb{E}\| \mv \Delta_i^{(t)}\|^2 \nonumber\\ 
        & \overset{(b)}{\le} (1-\lambda)(1+c)  \mathbb{E}\| \mv m_i^{(t)}\|^2 + (1-\lambda)(1+\frac{1}{c})\eta^2Q\sum_{q=0}^{Q-1}\mathbb{E}\| \hat{\nabla} F_i(\mv \theta_i^{(t,q)}) \|^2, \nonumber\\ 
        & \overset{(c)}{\le} (1-\lambda)(1+c)  \mathbb{E}\| \mv m_i^{(t)}\|^2  + (1-\lambda)(1+\frac{1}{c})\underbrace{\eta^2Q^2 \left(2(1+\alpha L_{G})^2 + 2\frac{\alpha^2 \sigma_H^2}{m_B}\right)G^2}_{\triangleq \tilde{G}^2} \nonumber\\ 
        & \le (1-\lambda)(1+\frac{1}{c}) \tilde{G}^2 \sum_{j=0}^{\infty} [(1-\lambda)(1+c)]^j.
\end{align}
where (a) is followed by Lemma~\ref{lemma:Youngs inequality} for any $\tau>0$, (b) is followed by Jensen's inequality and (c) is followed by Lemma~\ref{lemma:bound of SGD F}. Let $0<c<\frac{\lambda}{1-\lambda}$ such that $(1-\lambda)(1+c)<1$ and 
\begin{equation}
    \sum_{j=0}^{\infty} [(1-\lambda)(1+c)]^j = \frac{1}{1-(1-\lambda)(1+c)},
\end{equation}
which yields the result by defining $\Lambda = \frac{(1-\lambda)(1+\frac{1}{c})}{1-(1-\lambda)(1+c)}$.
 \end{IEEEproof}

 \begin{lemma} \label{lemma:power scaling factor bound}
    Considering the power constraint $(\Myfrac{1}{M}) \mathbb{E}\| (\Myfrac{\sqrt{\rho^{(t)}}}{\eta}) \boldsymbol g_{i}^{(t)} \|^2 \le P_{i}$, we have, for $t\in\{0,1,\ldots,T-1\}$,
    \begin{equation*}
        \frac{1}{\rho^{(t)}} \le \frac{4Q^2 G^2 (\Lambda + 1)}{M P_{\min}} \left( (1+\alpha L_{G})^2 + \frac{\alpha^2 \sigma_H^2}{m_B} \right).
    \end{equation*}
 \end{lemma}
 \begin{IEEEproof}
    Note that $\mathbb{E}\| \frac{\sqrt{\rho^{(t)}}}{\eta} \boldsymbol g_{i}^{(t)} \|^2 = \frac{\rho^{(t)}}{\eta^2}\mathbb{E}\|  \boldsymbol g_{i}^{(t)} \|^2 \le M P_i $. Define $$\rho = \min_{t}\min_{i} \frac{\eta^2 M P_i}{\mathbb{E}\| \mv g_i^{(t)} \|^2} $$ such that $\rho_t > \rho$. Define $P_{\min}=\min_{i\in[n]} P_i$. Then, we have 
    \begin{align} 
            \frac{1}{\rho} \nonumber
            & = \max_{t} \max_{i} \frac{\mathbb{E}\| \mv g_i^{(t)} \|^2}{\eta^2 M P_i} \nonumber\\ 
            & \overset{(a)}{\le} \max_{t} \max_{i} \frac{\mathbb{E}\| \mv m_i^{(t)} + \Delta_i^{(t)} \|^2}{\eta^2 M P_i} \nonumber\\ 
            & \le \frac{2}{\eta^2 M P_{\min}} \max_{t,i} \left( \mathbb{E}\| \mv m_i^{(t)} \|^2 + \mathbb{E} \| \mv \Delta_i^{(t)} \|^2 \right) \nonumber\\ 
            & \le \frac{2}{\eta^2 M P_{\min}} \max_{t,i} \left( \mathbb{E}\| \mv m_i^{(t)} \|^2 + Q \eta^2 \sum_{q=0}^{Q-1} \mathbb{E} \| \hat{\nabla} F_i(\mv \theta_i^{(t,q)}) \|^2 \right) \nonumber\\ 
            & \overset{(b)}{\le} \frac{4Q^2 G^2 (\Lambda + 1)}{M P_{\min}} \left( (1+\alpha L_{G})^2 + \frac{\alpha^2 \sigma_H^2}{m_B} \right),
    \end{align}
    where (a) is due to the fact that $\|\text{Comp}_k(\mv x)\|^2 \le \| \mv x \|^2$, and (b) is followed by Lemma~\ref{lemma:memory bound} and \ref{lemma:bound of SGD F}. The above inequality yields the desired result. 
 \end{IEEEproof}

\subsection{\textcolor{black}{Proofs of Theorem~\ref{theorem:convergence} and Theorem~\ref{theorem:convergence adaptive}}}~\label{appendix:proof of convergence}
The proof of Theorem~\ref{theorem:convergence} is based on the perturbed iterate analysis as in~\cite{stich2018sparsified}. To this end, we define the maintained virtual sequence $\{\hat{\mv \theta}^{(t)}\}_{t=0,\ldots,T}$ as follows:
\begin{equation}
    \hat{\mv \theta}^{(t+1)} = \hat{\mv \theta}^{(t)} - \frac{1}{rn}\sum_{i\in\mathcal{I}^{(t)}} \mv \Delta_{i}^{(t)} - \left(\frac{\eta^{(t)}}{\textcolor{black}{\mu_h} rn\sqrt{\rho^{(t)}}}\mv n_{\text{est}}^{(t)} + \textcolor{black}{\frac{1}{rn}\sum_{i\in\mathcal{I}^{(t)}}\left(\frac{|h_i^{(t)}|}{\mu_h}-1\right)\mv  g_i^{(t)}}\right),
\end{equation}
where $\hat{\mv \theta}^{(0)}=\mv \theta^{(0)}$. The relation between the true sequence \( \{\mv \theta^{(t)}\}_{t=0,\ldots,T} \) and the virtual sequence $\{\hat{\mv \theta}^{(t)}\}_{t=0,\ldots,T}$ is given by the following Lemma~\ref{lemma:virtual sequence relation}.

\begin{lemma}~\label{lemma:virtual sequence relation}
    The true sequence \( \{\mv \theta^{(t)}\}_{t=0,\ldots,T} \) and the virtual sequence $\{\tilde{\mv \theta}^{(t)}\}_{t=0,\ldots,T}$ satisfy the following equality: 
    $\mv \theta^{(t)} - \hat{\mv \theta}^{(t)} = (\Myfrac{1}{rn})\sum_{i=1}^{n}\mv m_i^{(t)},$
    where the expectation is over the randomness of channel gain and device selection.
\end{lemma}
\begin{IEEEproof}
    Following the Assumption~\ref{assumption:Estimation Error} and the global update \eqref{eq:over-the-air global updatas}, we have
    \begin{align}
        & \mv \theta^{(t+1)} - \hat{\mv \theta}^{(t+1)} \nonumber\\ 
        & =  \mv \theta^{(t)}-\frac{1}{rn}\sum_{i\in\mathcal{I}^{(t)}} \frac{|h_t^{(i)}|}{\mu_h} \mv g_{i}^{(t)} - \frac{\eta^{(t)}}{\mu_hrn\sqrt{\rho^{(t)}}}\mv n_{\text{est}}^{(t)} - \hat{\mv \theta}^{(t)} + \frac{1}{rn}\sum_{i\in\mathcal{I}^{(t)}} \mv \Delta_{i}^{(t)} + \frac{\eta^{(t)}}{\mu_hrn\sqrt{\rho^{(t)}}}\mv n_{\text{est}}^{(t)} \nonumber \\ 
        & \quad + \frac{1}{rn}\sum_{i\in\mathcal{I}^{(t)}}\left(\frac{|h_i^{(t)}|}{\mu_h}-1\right)\mv g_i^{(t)}  \nonumber\\ 
        & \overset{(a)}{=} \mv \theta^{(t)} - \hat{\mv \theta}^{(t)} +\frac{1}{rn}\sum_{i\in\mathcal{I}^{(t)}} (\mv \Delta_{i}^{(t)} - \mv g_{i}^{(t)}) \nonumber\\ 
        & \overset{(b)}{=} \mv \theta^{(t)} - \hat{\mv \theta}^{(t)} +\frac{1}{rn}\sum_{i\in\mathcal{I}^{(t)}} (\mv m_{i}^{(t+1)} - \mv m_{i}^{(t)}) \nonumber\\ 
        & \overset{(c)}{=} \mv \theta^{(t)} - \hat{\mv \theta}^{(t)} + \frac{1}{rn}\sum_{i=1}^{n}  (\mv m_{i}^{(t+1)} - \mv m_{i}^{(t)}) \nonumber\\ 
        & = \frac{1}{rn}\sum_{i=1}^{n}\sum_{j=0}^{t}(\mv m_{i}^{(j+1)} - \mv m_{i}^{(j)}) \nonumber\\ 
        & \overset{(d)}{=} \frac{1}{rn}\sum_{i=1}^{n}\mv m_{i}^{(t+1)},
    \end{align}
    where (a) is followed by $\frac{|h_t^{(i)}|}{\mu_h} \mv g_{i}^{(t)}=\mv g_{i}^{(t)}+(\frac{|h_t^{(i)}|}{\mu_h}-1) \mv g_{i}^{(t)}$; (b) is followed by memory update rule $  \mv{m}_{i}^{(t+1)}= \mv{m}_{i}^{(t)}+ \left(\mv{\Delta}_{i}^{(t)} - \mv g_i^{(t)}\right)$; (c) is followed by the fact that if $i\notin \mathcal{I}^{(t)}$, the memory will not be updated, i.e., $\mv m_{i}^{(t+1)} - \mv m_{i}^{(t)}=\mv 0$ for $i \notin \mathcal{I}^{(t)}$; (d) is directly obtained by telescope sum.
\end{IEEEproof}

Recording the definition of global loss function $F(\mv \theta)$ and the global update rule, we consider the perturbed sequence and begin the derivation by Lemma~\ref{lemma:L-smooth of F}, i.e., for any $t = 0,1,\ldots,T-1$, we have
\begin{multline} \label{eq:L-smooth inequality}
    \mathbb{E}_{n}\left[F(\hat{\mv \theta}^{(t+1)})\right] \le F(\hat{\mv \theta}^{(t)}) - \left< \nabla F(\hat{\mv \theta}^{(t)}), \frac{1}{rn} \sum_{i\in \mathcal{I}^{(t)}}\mv \Delta_i^{(t)}+\textcolor{black}{\frac{1}{rn}\sum_{i\in\mathcal{I}^{(t)}}\left(\frac{|h_i^{(t)}|}{\mu_h}-1\right)\mv  g_i^{(t)}}\right> \\
    + \frac{L_F}{2} \mathbb{E}\left\| \frac{1}{rn}\sum_{i\in\mathcal{I}^{(t)}} \mv \Delta_{i}^{(t)} + \left(\frac{\eta^{(t)}}{\textcolor{black}{\mu_h} rn\sqrt{\rho^{(t)}}}\mv n_{\text{est}}^{(t)} + \textcolor{black}{\frac{1}{rn}\sum_{i\in\mathcal{I}^{(t)}}\left(\frac{|h_i^{(t)}|}{\mu_h}-1\right)\mv  g_i^{(t)}}\right) \right\|^2
\end{multline}
Taking expectation on \eqref{eq:L-smooth inequality} over the estimate error $\mv n_{\text{est}}^{(t)}$ and following Assumption~\ref{assumption:Estimation Error} yield
\begin{multline} \label{eq:L-smooth inequality after expectation}
    \mathbb{E}_{n}\left[F(\hat{\mv \theta}^{(t+1)})\right] \le F(\hat{\mv \theta}^{(t)}) - \left< \nabla F(\hat{\mv \theta}^{(t)}), \frac{1}{rn} \sum_{i\in \mathcal{I}^{(t)}}\mv \Delta_i^{(t)}+\textcolor{black}{\frac{1}{rn}\sum_{i\in\mathcal{I}^{(t)}}\left(\frac{|h_i^{(t)}|}{\mu_h}-1\right)\mv  g_i^{(t)}}\right> \\
    + \frac{L_F}{2} \mathbb{E} \left\| \frac{1}{rn}\sum_{i\in\mathcal{I}^{(t)}} \mv \Delta_{i}^{(t)} +\textcolor{black}{\frac{1}{rn}\sum_{i\in\mathcal{I}^{(t)}}\left(\frac{|h_i^{(t)}|}{\mu_h}-1\right)\mv  g_i^{(t)}} \right\|^2  + \frac{L_F \eta^2}{2\rho^{(t)}}\frac{d v^{(t)}}{\textcolor{black}{\mu_h^2} r^2n^2}
\end{multline}
In the following, our target is to bound the inner-product term and the square norm term. First, we consider the inner-product term and take expectation over SGD, device sampling, random sparsification, and channel fading conditioned on time $t$, yielding
\begin{align} \label{eq:inner products intermediate result}
    & -\mathbb{E}\left< \nabla F(\hat{\mv \theta}^{(t)}), \frac{1}{rn} \sum_{i\in \mathcal{I}^{(t)}}\mv \Delta_i^{(t)}+\textcolor{black}{\frac{1}{rn}\sum_{i\in\mathcal{I}^{(t)}}\left(\frac{|h_i^{(t)}|}{\mu_h}-1\right)\mv  g_i^{(t)}}\right> \nonumber\\ 
    \overset{(a)}{=} & -\left< \nabla F(\hat{\mv \theta}^{(t)}), \frac{\eta}{n} \sum_{i=1}^{n}\mathbb{E}\left[ \sum_{q=0}^{Q-1}\hat{\nabla}F_i(\mv \theta_{i}^{(t,q)}) \right]+\textcolor{black}{\frac{1}{n}\sum_{i=1}^{n}\left(\frac{\mu_h}{\mu_h}-1\right)\mathbb{E}\left[\mv g_i^{(t)}\right]}\right> \nonumber\\ 
    = & -\Bigg< \nabla F(\hat{\mv \theta}^{(t)}), \frac{\eta }{n} \sum_{i=1}^{n}\mathbb{E}\Bigg[ \sum_{q=0}^{Q-1}\hat{\nabla}F_i(\mv \theta_{i}^{(t,q)}) \Bigg]\Bigg> \nonumber\\ 
    = & -\Bigg< \nabla F(\hat{\mv \theta}^{(t)}),  \frac{\eta }{n} \sum_{i=1}^{n}\mathbb{E}\Bigg[ \sum_{q=0}^{Q-1}\hat{\nabla}F_i(\mv \theta_{i}^{(t,q)}) - Q\nabla F_i(\hat{\mv \theta}^{(t)})  + Q\nabla F_i(\hat{\mv \theta}^{(t)}) \Bigg]\Bigg> \nonumber\\ 
    \overset{(b)}{=} & -\eta  Q \| \nabla F(\hat{\mv \theta}^{(t)}) \|^2  - \eta  \left< \nabla F(\hat{\mv \theta}^{(t)}), \frac{1}{n} \sum_{i=1}^{n}\mathbb{E}\left[ \sum_{q=0}^{Q-1}\hat{\nabla}F_i(\mv \theta_{i}^{(t,q)}) - Q\nabla F_i(\hat{\mv \theta}^{(t)}) \right] \right> \nonumber\\ 
    \overset{(c)}{\le} & - \frac{\eta  Q}{2}\| \nabla F(\hat{\mv \theta}^{(t)}) \|^2  + \frac{\eta}{2Q} \left\| \frac{1}{n}\sum_{i=1}^{n}\mathbb{E}\left[ \sum_{q=0}^{Q-1}\hat{\nabla}F_i(\mv \theta_{i}^{(t,q)}) - Q\nabla F_i(\hat{\mv \theta}^{(t)}) \right] \right\|^2
\end{align}
where (a) is followed by tower rule and Assumption~\ref{assumption:channel fading}; (b) is due to the definition of $F({\mv \theta}) \triangleq \Myfrac{1}{n}\sum_{i=1}^{n}F_i({\mv \theta}) $, and (c) is followed by Young's inequality (c.f.~Lemma~\ref{lemma:Youngs inequality}). Then we bound the last term in \eqref{eq:inner products intermediate result} as follows:
\begin{align} \label{eq:bound of square norm of last term in inner product}
    & \left\| \frac{1}{n}\sum_{i=1}^{n}\mathbb{E}\left[ \sum_{q=0}^{Q-1}\hat{\nabla}F_i(\mv \theta_{i}^{(t,q)}) - Q\nabla F_i(\hat{\mv \theta}^{(t)}) \right] \right\|^2 \nonumber\\ 
    \overset{(a)}{\le} & \frac{1}{n}\sum_{i=1}^{n} \bigg\| \mathbb{E}\bigg[ \sum_{q=0}^{Q-1}\Big(\hat{\nabla}F_i(\mv \theta_{i}^{(t,q)}) - \nabla F_i(\mv \theta_{i}^{(t,q)}) + \nabla F_i(\mv \theta_{i}^{(t,q)})- \nabla F_i(\mv \theta^{(t)}) + \nabla F_i(\mv \theta^{(t)})  - \nabla F_i(\hat{\mv \theta}^{(t)})\Big) \bigg] \bigg\|^2 \nonumber\\ 
    \overset{(b)}{\le} & \frac{1}{n}\sum_{i=1}^{n}\bigg( 3\bigg\| \mathbb{E}\bigg[ \sum_{q=0}^{Q-1}\left(\hat{\nabla}F_i(\mv \theta_{i}^{(t,q)}) - \nabla F_i(\mv \theta_{i}^{(t,q)}) \right) \bigg] \bigg\|^2 \nonumber\\
    & \hspace{0.5in} + 3\left\| \mathbb{E}\left[ \sum_{q=0}^{Q-1}\left( \nabla F_i(\mv \theta_{i}^{(t,q)}) - \nabla F_i(\mv \theta^{(t)}) \right) \right] \right\|^2 \nonumber\\ 
    & \hspace{0.5in} + 3\left\| \mathbb{E}\left[ \sum_{q=0}^{Q-1}\left( \nabla F_i(\mv \theta^{(t)}) - \nabla F_i(\hat{\mv \theta}^{(t)})\right) \right] \right\|^2 \bigg) \nonumber\\ 
    \overset{(c)}{\le} & \frac{1}{n}\sum_{i=1}^{n}\bigg( \frac{12\alpha^2Q^2L^2\sigma_G^2}{m_B} + 3L_F^2 Q\sum_{q=0}^{Q-1}\mathbb{E}\| \mv \theta^{(t)} - \mv \theta^{(t,q)}_i \|^2 + 3L_F^2Q^2\mathbb{E}\|\mv \theta^{(t)} - \hat{\mv \theta}^{(t)}\|^2 \bigg)
\end{align}
where (a) and (b) are followed by Jensen's inequality and (c) is followed by Lemma~\ref{lemma:L-smooth of F}, Lemma~\ref{lemma:Bounded variance of F}, and Jensen's inequality. The relation between $\| \nabla F(\hat{\mv \theta}^{(t)}) \|^2$ and $\| \nabla F(\mv \theta^{(t)}) \|^2$ is given by 
\begin{align} \label{eq:relation between square norm of gradient sequences}
    \| \nabla F(\mv \theta^{(t)}) \|^2 
    & \le 2 \|\nabla F(\mv \theta^{(t)}) - \nabla F(\hat{\mv \theta}^{(t)}) \|^2 + 2\| \nabla F(\hat{\mv \theta}^{(t)}) \|^2 \nonumber\\
    & \le 2L_F^2\|\mv \theta^{(t)} - \hat{\mv \theta}^{(t)}\|^2 +  2\| \nabla F(\hat{\mv \theta}^{(t)}) \|^2.
\end{align}
Combining \eqref{eq:inner products intermediate result} with \eqref{eq:bound of square norm of last term in inner product} and \eqref{eq:relation between square norm of gradient sequences} yields
\begin{align} \label{eq:inner products bound}
    & -\left< \nabla F(\hat{\mv \theta}^{(t)}), \frac{\eta}{n} \sum_{i=1}^{n}\mathbb{E}\left[ \sum_{q=0}^{Q-1}\hat{\nabla}F_i(\mv \theta_{i}^{(t,q)}) \right]\right> \nonumber\\ 
    \le & - \frac{\eta Q}{4}\| \nabla F(\mv \theta^{(t)}) \|^2 + \frac{6\eta \alpha^2QL^2\sigma_G^2}{m_B}  + \frac{3}{2}\eta L_F^2 \frac{1}{n}\sum_{i=1}^{n}\sum_{q=0}^{Q-1}\mathbb{E}\| \mv \theta^{(t)} - \mv \theta^{(t,q)}_i \|^2  + 2\eta Q L_F^2 \mathbb{E}\|\mv \theta^{(t)} - \hat{\mv \theta}^{(t)}\|^2 \nonumber\\ 
    \overset{(a)}{=} & - \frac{\eta Q}{4}\| \nabla F(\mv \theta^{(t)}) \|^2 + \frac{6\eta  \alpha^2QL^2\sigma_G^2}{m_B} + \frac{3}{2}\eta L_F^2 \frac{1}{n}\sum_{i=1}^{n}\sum_{q=0}^{Q-1}\mathbb{E}\| \mv \theta^{(t)} - \mv \theta^{(t,q)}_i \|^2 \textcolor{black}{+2\eta Q L_F^2 \mathbb{E}\left\| \frac{1}{rn}\sum_{i=1}^{n}\mv m_i^{(t)} \right\|^2} \nonumber\\
    \overset{(b)}{\le} & - \frac{\eta Q}{4}\| \nabla F(\mv \theta^{(t)}) \|^2 + \frac{6\eta \alpha^2QL^2\sigma_G^2}{m_B} + \textcolor{black}{\frac{4\eta^3 Q^3 L_F^2 \Lambda}{r^2} \left( (1+\alpha L_{G})^2 + \frac{\alpha^2 \sigma_H^2}{m_B} \right)G^2} \nonumber\\ 
    & \quad + \frac{3}{2}\eta^3 Q^3 L_F^2(40(\sigma_F^2 + \gamma_F^2)+40\mathbb{E}\| \nabla F(\mv \theta^{(t)})\|^2 ),
\end{align}
where (a) is followed by Lemma~\ref{lemma:virtual sequence relation} and (b) is followed by Jensen inequality, Lemma~\ref{lemma:memory bound} and Lemma~\ref{lemma:difference between local and global theta}. 

Then, we consider the square norm term in~\eqref{eq:L-smooth inequality after expectation} by taking expectation over SGD, device sampling, random sparsification, and channel fading conditioned on time $t$, i.e.,
\begin{align}
    & \frac{L_F}{2} \mathbb{E}\left\| \frac{1}{rn} \sum_{i\in \mathcal{I}^{(t)}}\mv \Delta_i^{(t)} +\textcolor{black}{\frac{1}{rn}\sum_{i\in\mathcal{I}^{(t)}}\left(\frac{|h_i^{(t)}|}{\mu_h}-1\right)\mv  g_i^{(t)}} \right\|^2 \nonumber\\ 
    \le &  L_F \mathbb{E} \left\| \frac{1}{rn} \sum_{i\in \mathcal{I}^{(t)}} \mv \Delta_i^{(t)} \right\|^2 + \textcolor{black}{L_F \mathbb{E}\left\|\frac{1}{rn}\sum_{i\in\mathcal{I}^{(t)}}\left(\frac{|h_i^{(t)}|}{\mu_h}-1\right)\mv  g_i^{(t)} \right\|^2 }
    \label{eq:intermediate results of l2-norm}
\end{align}
Then we bound the first term in \eqref{eq:intermediate results of l2-norm} as follow:
\begin{align} 
        & {L_F} \mathbb{E}\left\| \frac{1}{rn} \sum_{i\in \mathcal{I}^{(t)}}\mv \Delta_i^{(t)} \right\|^2 \nonumber\\ 
        = & {\eta^2 L_F} \mathbb{E} \left\| \frac{1}{rn} \sum_{i\in \mathcal{I}^{(t)}}  \sum_{q=0}^{Q-1}\big( \hat{\nabla}F_i(\mv \theta_i^{(t,q)}) \big) \right\|^2 \nonumber\\ 
        \overset{(a)}{\le} & {\eta^2 Q L_F} \sum_{q=0}^{Q-1} \mathbb{E} \left\| \frac{1}{rn} \sum_{i\in \mathcal{I}^{(t)}}  \hat{\nabla}F_i(\mv \theta_i^{(t,q)})  \right\|^2 \nonumber\\ 
        = & {\eta^2 Q L_F} \sum_{q=0}^{Q-1} \mathbb{E} \bigg\| \frac{1}{rn} \sum_{i\in \mathcal{I}^{(t)}}   \big( \hat{\nabla}F_i(\mv \theta_i^{(t,q)}) - \nabla F_i(\mv \theta_i^{(t,q)}) + \nabla F_i(\mv \theta_i^{(t,q)}) - \nabla F_i(\mv \theta^{(t)}) \nonumber \\ 
        & \quad + \nabla F_i(\mv \theta^{(t)}) - \nabla F(\mv \theta^{(t)}) + \nabla F(\mv \theta^{(t)}) \big) \bigg\|^2 \nonumber\\ 
        \le & {\eta^2 Q L_F}  \sum_{q=0}^{Q-1} \bigg( 4\mathbb{E} \bigg\| \frac{1}{rn} \sum_{i\in \mathcal{I}^{(t)}}   \big( \hat{\nabla}F_i(\mv \theta_i^{(t,q)}) - \nabla F_i(\mv \theta_i^{(t,q)})\big) \bigg\|^2 + 4 \mathbb{E}  \bigg\| \frac{1}{rn} \sum_{i\in \mathcal{I}^{(t)}}  \nabla F(\mv \theta^{(t)}) \bigg\|^2 \nonumber\\ 
        & ~ + 4\mathbb{E}  \bigg\| \frac{1}{rn} \sum_{i\in \mathcal{I}^{(t)}}   \big( \nabla F_i(\mv \theta_i^{(t,q)}) - \nabla F_i(\mv \theta^{(t)}) \big) \bigg\|^2 + 4\mathbb{E}  \bigg\| \frac{1}{rn} \sum_{i\in \mathcal{I}^{(t)}} \big( \nabla F_i(\mv \theta^{(t)}) - \nabla F(\mv \theta^{(t)}) \big) \bigg\|^2 \bigg) 
\end{align}
where (a) is due to the sum is changeable and followed by Jensen's inequality. We bound the first three square $l_2$-norm terms as follows.
\begin{align}
    & \mathbb{E} \bigg\| \frac{1}{rn} \sum_{i\in \mathcal{I}^{(t)}}  \big( \hat{\nabla}F_i(\mv \theta_i^{(t,q)}) - \nabla F_i(\mv \theta_i^{(t,q)})\big) \bigg\|^2 \nonumber\\ 
    \overset{(a)}{\le} &  \mathbb{E} \left[ \frac{1}{rn}\sum_{i\in \mathcal{I}^{(t)}} \bigg\|  \big( \hat{\nabla}F_i(\mv \theta_i^{(t,q)}) - \nabla F_i(\mv \theta_i^{(t,q)})\big) \bigg\|^2 \right] \nonumber\\ 
    \overset{(b)}{=} & \frac{1}{n}\sum_{i=1}^{n} \mathbb{E} \bigg\|  \big( \hat{\nabla}F_i(\mv \theta_i^{(t,q)}) - \nabla F_i(\mv \theta_i^{(t,q)})\big) \bigg\|^2 \nonumber\\ 
    \overset{(c)}{\le} & \sigma_F^2,
\end{align}
where (a), (b), and (c) are followed by Jensen's inequality, tower rule and Lemma~\ref{lemma:Bounded variance of F}, respectively. Similarly, 
\begin{align}
        & \mathbb{E}  \bigg\| \frac{1}{rn} \sum_{i\in \mathcal{I}^{(t)}}  \big( \nabla F_i(\mv \theta_i^{(t,q)}) - \nabla F_i(\mv \theta^{(t)}) \big) \bigg\|^2 \nonumber\\ 
        \le & \frac{1}{n}\sum_{i=1}^{n} \mathbb{E} \bigg\|  \nabla F_i(\mv \theta_i^{(t,q)}) - \nabla F_i(\mv \theta^{(t)}) \bigg\|^2 \nonumber\\ 
        \overset{(a)}{\le} &  \frac{ L_F^2}{n}\sum_{i=1}^{n} \mathbb{E} \bigg\|  \mv \theta_i^{(t,q)} - \mv \theta^{(t)} \bigg\|^2 \nonumber\\ 
        \overset{(b)}{\le} & L_F^2 (40 Q^2 \eta^2 (\sigma_F^2 + \gamma_F^2) + 40Q^2 \eta^2 \mathbb{E}\| \nabla F(\mv \theta^{(t)})\|^2 )
\end{align}
where (a) and (b) are followed Lemma~\ref{lemma:L-smooth of F} and \ref{lemma:difference between local and global theta}. For the third $l_2$-norm term, we bound it as follows:
\begin{align}
        & \mathbb{E}  \bigg\| \frac{1}{rn} \sum_{i\in \mathcal{I}^{(t)}} \big( \nabla F_i(\mv \theta^{(t)}) - \nabla F(\mv \theta^{(t)}) \big) \bigg\|^2 \nonumber\\ 
        \le & \frac{1}{rn} \sum_{i\in \mathcal{I}^{(t)}}  \mathbb{E} \left\|\big( \nabla F_i(\mv \theta^{(t)}) - \nabla F(\mv \theta^{(t)}) \big) \right\|^2 \nonumber \\ 
        \overset{(a)}{\le} &  \gamma_F^2 
\end{align}
where (a) follows Lemma~\ref{lemma:Bounded Heterogeneity of F}. Therefore, we have the following bound:
\begin{align} \label{eq:square norm term bound}
    & L_F \mathbb{E}\left\| \frac{1}{rn} \sum_{i\in \mathcal{I}^{(t)}}\mv \Delta_i^{(t)} \right\|^2 \nonumber\\ 
    \le & 4\eta^2Q^2L_F\bigg( \sigma_F^2 +  L_F^2 (40 Q^2 \eta^2 (\sigma_F^2 + \gamma_F^2) + 40Q^2 \eta^2 \mathbb{E}\| \nabla F(\mv \theta^{(t)})\|^2 ) + \gamma_F^2 + \|\nabla F(\mv \theta^{(t)}) \|^2 \bigg) 
\end{align}

The second term in \eqref{eq:intermediate results of l2-norm} can be bounded as 
\begin{align} \label{eq:bound of l2-norm of g}
    & \textcolor{black}{L_F \mathbb{E}\left\|\frac{1}{rn}\sum_{i\in\mathcal{I}^{(t)}}\left(\frac{|h_i^{(t)}|}{\mu_h}-1\right)\mv  g_i^{(t)} \right\|^2 }\nonumber\\ 
    \overset{(a)}{\le} &  \textcolor{black}{L_F \mathbb{E}\left[\frac{1}{rn}\sum_{i\in\mathcal{I}^{(t)}}\left\|\left(\frac{|h_i^{(t)}|}{\mu_h}-1\right)\mv  g_i^{(t)} \right\|^2 \right]} \nonumber\\ 
    \overset{(b)}{=} & \textcolor{black}{L_F\frac{1}{n}\sum_{i=1}^{n} \mathbb{E}\left\|\left(\frac{|h_i^{(t)}|}{\mu_h}-1\right)\mv  g_i^{(t)} \right\|^2} \nonumber\\ 
    \overset{(c)}{\le } & \textcolor{black}{4\eta^2L_F Q^2 G^2 (\Lambda + 1)\left( (1+\alpha L_{G})^2 + \frac{\alpha^2 \sigma_H^2}{m_B} \right) \mathbb{E}\left[\left(\frac{|h_i^{(t)}|}{\mu_h}-1\right)^2\right] }  \nonumber\\ 
    = & \textcolor{black}{4\eta^2L_F Q^2 G^2 (\Lambda + 1)\left( (1+\alpha L_{G})^2 + \frac{\alpha^2 \sigma_H^2}{m_B} \right) \left(\frac{\sigma_h^2}{\mu_h^2}-1\right) },
\end{align}
where (a), (b) and (c) are followed by Jensen inequality, tower rule and the bound of $\mathbb{E}\| \mv g_i^{(t)} \|^2$ obtained by the same technique used in the proof of Lemma~\ref{lemma:power scaling factor bound}, respectively.

Substituting \eqref{eq:inner products bound}, \eqref{eq:square norm term bound} and \eqref{eq:bound of l2-norm of g} into \eqref{eq:L-smooth inequality after expectation} by taking expectation, we have the following per-round convergence bound:
\begin{multline}
    \mathbb{E}_{n}\left[F(\hat{\mv \theta}^{(t+1)})\right] \le F(\hat{\mv \theta}^{(t)})  - \frac{\eta Q}{4}\| \nabla F(\mv \theta^{(t)}) \|^2 + \frac{6\eta \alpha^2QL^2\sigma_G^2}{m_B} + \textcolor{black}{\frac{4\eta^3 Q^3 L_F^2 \Lambda}{r^2} \left( (1+\alpha L_{G})^2 + \frac{\alpha^2 \sigma_H^2}{m_B} \right)G^2} \nonumber\\ 
        \quad + \frac{3}{2}\eta^3 Q^3 L_F^2\left(40(\sigma_F^2 + \gamma_F^2)+40\mathbb{E}\| \nabla F(\mv \theta^{(t)})\|^2 \right) \\ 
    + 4\eta^2Q^2L_F\bigg( \sigma_F^2 +  L_F^2 (40 Q^2 \eta^2 (\sigma_F^2 + \gamma_F^2) + 40Q^2 \eta^2 \mathbb{E}\| \nabla F(\mv \theta^{(t)})\|^2 ) + \gamma_F^2 + \mathbb{E}  \|\nabla F(\mv \theta^{(t)}) \|^2 \bigg) \\ 
    +\textcolor{black}{4\eta^2L_F Q^2 G^2 (\Lambda + 1)\left( (1+\alpha L_{G})^2 + \frac{\alpha^2 \sigma_H^2}{m_B} \right) \left(\frac{\sigma_h^2}{\mu_h^2}-1\right) }  + \frac{L_F \eta^2}{2\rho^{(t)}}\frac{d v^{(t)}}{r^2n^2}
\end{multline}
Let $\eta$ satisfy the following inequality
\begin{equation}
    60\eta^2Q^2L_F^2+160\eta^3Q^3L_F^3+4\eta QL_F \le \frac{1}{8},
\end{equation}
which simplifies the inequality as follows
\begin{multline} \label{eq:per-round convergence bound}
    \mathbb{E}_{n}\left[F(\hat{\mv \theta}^{(t+1)})\right] \le F(\hat{\mv \theta}^{(t)})  - \frac{\eta Q}{4}\| \nabla F(\mv \theta^{(t)}) \|^2 + \frac{6\eta \alpha^2QL^2\sigma_G^2}{m_B} + \textcolor{black}{\frac{4\eta^3 Q^3 L_F^2 \Lambda}{r^2} \left( (1+\alpha L_{G})^2 + \frac{\alpha^2 \sigma_H^2}{m_B} \right)G^2} \\ 
    \quad + 60\eta^3 Q^3 L_F^2(\sigma_F^2 + \gamma_F^2)  + 4\eta^2Q^2L_F\bigg(\sigma_F^2 + L_F^2 \left(40 Q^2 \eta^2 (\sigma_F^2 + \gamma_F^2)  \right) +\gamma_F^2 \bigg) \\ 
    +\textcolor{black}{4\eta^2L_F Q^2 G^2 (\Lambda + 1)\left( (1+\alpha L_{G})^2 + \frac{\alpha^2 \sigma_H^2}{m_B} \right) \left(\frac{\sigma_h^2}{\mu_h^2}-1\right) }  + \frac{L_F \eta^2}{2\rho^{(t)}}\frac{d v^{(t)}}{r^2n^2}
\end{multline}

By Lemma~\ref{lemma:power scaling factor bound}, rearranging, taking telescope sum and taking expectation over all randomness, we obtain
\begin{multline}
    \frac{1}{T}\sum_{t=0}^{T-1}\mathbb{E}\|\nabla F(\mv \theta^{(t)})\|^2 \le \frac{8}{\eta Q T}\left( F(\mv \theta^{(0)}) - F^* \right)+\frac{48\alpha^2L^2\sigma_G^2}{m_B}  \\ 
    + \textcolor{black}{\frac{32\eta^2 Q^2 L_F^2 \Lambda}{r^2} \left( (1+\alpha L_{G})^2 + \frac{\alpha^2 \sigma_H^2}{m_B} \right)G^2}  + 480\eta^2 Q^2 L_F^2(\sigma_F^2 + \gamma_F^2) \\ 
    + 32\eta Q L_F\bigg(\sigma_F^2 + L_F^2 \left(40 Q^2 \eta^2 (\sigma_F^2 + \gamma_F^2) \right) + \gamma_F^2 \bigg) \\ 
    + \textcolor{black}{32\eta L_F Q G^2 (\Lambda + 1)\left( (1+\alpha L_{G})^2 + \frac{\alpha^2 \sigma_H^2}{m_B} \right) \left(\frac{\sigma_h^2}{\mu_h^2}-1\right) } \\  
    + \frac{16\eta L_F d Q G^2 (\Lambda + 1)}{r^2n^2M P_{\min}} \left( (1+\alpha L_{G})^2 + \frac{\alpha^2 \sigma_H^2}{m_B} \right)\frac{1}{T}\sum_{t=0}^{T-1} v^{(t)},
    \end{multline}
where we use $\hat{\mv \theta}^{(0)}=\mv \theta^{(0)}$, $F^* = \min_{\boldsymbol \theta \in \mathbb{R}^{d}} F(\mv \theta)$, and $\eta \le 1/L_F$ to obtain the desired result of Theorem~\ref{theorem:convergence}.

The proof of Theorem~\ref{theorem:convergence adaptive} is analogous to that of Theorem~\ref{theorem:convergence} in principal. It is worth noting that we should carefully design the parameters of $\alpha^{(t)}$ and $\eta^{(t)}$ such that we can arrive at a first-order regression inequality similar to \eqref{eq:per-round convergence bound}. In addition, we modify the memory bound for adaptive learning rate derived in ~\cite[Lemma 4]{basu19qsparse}. Next, we take a telescopic sum, and use the inequalities
$ \sum_{t=0}^{T-1}\eta^{(t)}\ge \xi \ln\left(\Myfrac{(T+a-1)}{a}\right),~ \sum_{t=0}^{T-1}(\eta^{(t)})^2\le \Myfrac{\xi^2}{(a-1)} $ and $ \sum_{t=0}^{T-1}(\eta^{(t)})^3\le \Myfrac{\xi^3}{2(a-1)^2},$
which completes the proof.

{\color{black}
\subsection{Proofs of Theorem~\ref{theorem:generalization Air-meta-pFL}} \label{appendix:proof of generalization Air-meta-pFL}

The effective global update of Air-meta-pFL is given by 
\begin{align} \label{eq:update rule of Air-meta-pFL}
    \mv \theta^{(t+1)}  = \mv \theta^{(t)}-\frac{1}{rn}\sum_{i\in\mathcal{I}^{(t)}} \frac{|h_i^{(t)}|}{\textcolor{black}{\mu_h}} \mv g_{i}^{(t)} - \frac{\eta^{(t)}}{\textcolor{black}{\mu_h}rn\sqrt{\rho^{(t)}}}\mv n_{\text{est}}^{(t)}
\end{align}
which is a noisy weighted-sum SGD update.
We first define the Markov chain as follows 
\begin{equation*} 
    \mathcal{D}_{1:n} \to \mathcal{D}_{\mathcal{I}_{[T]}} \to \mathcal{B}_{\mathcal{I}_{[T]}} \to \mv \theta_{[T]} \to \mv \theta,
\end{equation*}
where $\mv \theta$ is short for the output of Air-meta-pFL, i.e., $\mv\theta^{(T)}$, for the ease of exhibition; $\mathcal{D}_{\mathcal{I}_{[T]}} = \{ \mathcal{D}_{\mathcal{I}_{(0)}}, \mathcal{D}_{\mathcal{I}_{(1)}}, \ldots, \mathcal{D}_{\mathcal{I}_{(T-1)}} \}$ with $\mathcal{D}_{\mathcal{I}_{(t)}}=\{\mathcal{D}_i\}_{i\in \mathcal{I}^{(t)}}$; $\mathcal{B}_{\mathcal{I}_{[T]}}$ is defined as a similar way; and $\mv \theta_{[T]} = \{\mv \theta^{(1)}, \mv \theta^{(2)}, \ldots, \mv \theta^{(T)}\}$. Based on this Markov Chain, we have the following inequalities by data processing inequality:
\begin{equation*}
    I(\mv \theta; \mathcal{D}_{1:n}) \le I(\mv \theta_{[T]}; \mathcal{D}_{1:n}) \le I(\mv \theta_{[T]}; \mathcal{D}_{\mathcal{I}_{[T]}})\le I(\mv \theta_{[T]}; \mathcal{B}_{\mathcal{I}_{[T]}}).
\end{equation*}
We calculate the term $I(\mv \theta_{[T]}; \mathcal{B}_{\mathcal{I}_{[T]}})$ as follows:
\begin{align}
        & I(\mv \theta_{[T]}; \mathcal{B}_{\mathcal{I}_{[T]}}) = \sum_{t=1}^{T}I(\mv \theta^{(t)}; \mathcal{B}_{\mathcal{I}_{[T]}} \mid \mv \theta_{[t-1]}) \nonumber \\ 
        & \overset{(a)}{=} \sum_{t=1}^{T}I(\mv \theta^{(t)}; \mathcal{B}_{\mathcal{I}^{(t)}} \mid \mv \theta^{(t-1)}) \nonumber \\ 
        & = \sum_{t=1}^{T}\left( h(\mv \theta^{(t)} \mid \mv \theta^{(t-1)}) - h(\mv \theta^{(t)}\mid \mv \theta^{(t-1)}, \mathcal{B}_{\mathcal{I}^{(t)}}) \right), \label{eq:MI intermediate bound on entropy}
\end{align}
where $h(x|y)$ is the entropy of random variable $x$ conditioned on $y$, the first equality is followed by the chain rule of mutual information, (a) is followed by the update rule in~\eqref{eq:update rule of Air-meta-pFL} and the assumption that the sampling strategy is independent of the parameters and the previous samplings. In order to bound the entropy $h(\mv \theta^{(t)} \mid \mv \theta^{(t-1)})$, note that conditioned on $\mv \theta^{(t-1)}=\mv \vartheta^{(t-1)}$, we have $h(\mv \theta^{(t)} \mid \mv \theta^{(t-1)} = \mv \vartheta^{(t-1)}) = h(\mv \theta^{(t)} - \mv \vartheta^{(t-1)} \mid \mv \theta^{(t-1)} = \mv \vartheta^{(t-1)})$. We next bound the second moment of the random variable $\mv \theta^{(t)} - \mv \vartheta^{(t-1)}$ as below.
\begin{equation}
    \begin{aligned}
        & \mathbb{E}\| \mv \theta^{(t)} - \mv \vartheta^{(t-1)} \|^2  \\
        = & \mathbb{E}\left\| \frac{1}{\mu_h rn} \sum_{i\in\mathcal{I}^{(t-1)}} |h_i^{(t-1)}| \mv g_i^{(t-1)} \right\|^2 + \frac{d \eta^2 v^{(t)} }{\mu_h^2 r^2n^2\rho^{(t-1)}} \\ 
        \overset{(a)}{\le} & \frac{1}{\mu_h^2 rn} \sum_{i\in\mathcal{I}^{(t-1)}}  |h_i^{(t-1)}|^2 \mathbb{E}\left\| \mv g_i^{(t-1)} \right\|^2 + \frac{d \eta^2 v^{(t)} }{\mu_h^2 r^2n^2\rho^{(t-1)}} \\ 
        \overset{(b)}{\le} & \frac{1}{\mu_h^2 rn} \sum_{i\in\mathcal{I}^{(t-1)}}  |h_i^{(t-1)}|^2 \mathbb{E}\left\| \mv m_i^{(t-1)}+\mv \Delta_i^{(t)} \right\|^2 + \frac{d \eta^2 v^{(t)} }{\mu_h^2 r^2n^2\rho^{(t-1)}} \\ 
        \le & \frac{1}{\mu_h^2 rn} \sum_{i\in\mathcal{I}^{(t-1)}}  |h_i^{(t-1)}|^2 \left(2\mathbb{E}\left\| \mv m_i^{(t-1)} \right\|^2+2\mathbb{E}\left\|\mv \Delta_i^{(t)} \right\|^2 \right) + \frac{d \eta^2 v^{(t)} }{\mu_h^2 r^2n^2\rho^{(t-1)}} \\
        \overset{(c)}{\le} & \underbrace{\frac{1}{\mu_h^2 rn}\left(4\eta^2 Q^2 G^2 (\Lambda + 1)\left( (1+\alpha L_{G})^2 + \frac{\alpha^2 \sigma_H^2}{m_B} \right) \right) \sum_{i\in\mathcal{I}^{(t-1)}}  |h_i^{(t-1)}|^2}_{C_1} + \underbrace{\frac{d \eta^2 v^{(t)} }{\mu_h^2 r^2n^2\rho^{(t-1)}}}_{C_2},
    \end{aligned}
\end{equation}
where (a) follows the convexity of $\|\cdot\|^2$, (b) is by the definition of $\mv g_i^{(t)}$, and (c) is as a result of Lemma \ref{lemma:bound of SGD F} and Lemma \ref{lemma:memory bound}.
Since $1/\rho^{(t)} < \infty$ by Lemma~\ref{lemma:power scaling factor bound}, $\mathbb{E}\| \mv \theta^{(t)} - \mv \vartheta^{(t-1)} \|^2$ is bounded, which leads to $ h(\mv \theta^{(t)} \mid \mv \theta^{(t-1)}=\mv \vartheta^{(t-1)})$ upper-bounded by the entropy of the Gaussian distribution with zero mean and variance $\sqrt{(C_1+C_2)/d}\mv I_d$, i.e., \(h(\mv \theta^{(t)} \mid \mv \theta^{(t-1)}=\mv \vartheta^{(t-1)}) \le \frac{d}{2}\log\left( \frac{2\pi e (C_1+C_2)}{d} \right)\). Since this bound holds for all values of $\mv \vartheta^{(t-1)}$, we conclude that \(h(\mv \theta^{(t)} \mid \mv \theta^{(t-1)}) \le \frac{d}{2}\log\left( \frac{2\pi e (C_1+C_2)}{d} \right) \).
Combining with
\begin{align*}
    h(\mv \theta^{(t)}\mid \mv \theta^{(t-1)}, \mathcal{B}_{\mathcal{I}^{(t-1)}}) 
    = &  h\left(\frac{\eta}{\mu_h rn\sqrt{\rho^{(t-1)}}}\mv n_{\text{est}}^{(t-1)}\right) \\ 
    = & \frac{d}{2}\log\left(2\pi e \frac{\eta^2 v^{(t)}}{\mu_h^2 r^2n^2\rho^{(t-1)}}\right),
\end{align*} we have 
\begin{equation}
    \begin{aligned}
        & \quad h(\mv \theta^{(t)} \mid \mv \theta^{(t-1)}) - h(\mv \theta^{(t)}\mid \mv \theta^{(t-1)}, \mathcal{B}_{\mathcal{I}^{(t)}}) \\ 
        & \le \frac{d}{2}\log\left( \frac{2\pi e (C_1+C_2)}{d} \right) - \frac{d}{2}\log\left(2\pi e \frac{\eta^2 v^{(t)}}{\mu_h^2 r^2n^2\rho^{(t-1)}}\right) \\
        & = \frac{d}{2}\log\left(1+ \frac{2\pi e}{d} \frac{\frac{1}{\mu_h^2 rn}\left(4\eta^2 Q^2 G^2 (\Lambda + 1)\left( (1+\alpha L_{G})^2 + \frac{\alpha^2 \sigma_H^2}{m_B} \right) \right) \sum_{i\in\mathcal{I}^{(t-1)}}  |h_i^{(t-1)}|^2}{2\pi e \frac{\eta^2 v^{(t)}}{\mu_h^2 r^2n^2\rho^{(t-1)}}} \right) \\
        & \le \frac{d}{2} \log\Bigg(1 + \frac{ M P_{\max}\left(4Q^2 G^2 (\Lambda + 1)\left( (1+\alpha L_{G})^2 + \frac{\alpha^2 \sigma_H^2}{m_B} \right) \right) rn \sum_{i\in\mathcal{I}^{(t-1)}}|h_i^{(t-1)}|^2}{d v^{(t)} \epsilon_g} \Bigg),
\end{aligned}
\end{equation}
where we use $\rho^{(t)} \le M P_{\max}/\epsilon_g$ with the assumption of $\mathbb{E}\| \mv g_i^{(t)} \|^2 \ge \eta^2 \epsilon_g$ and define $P_{\max}=\max_{i\in[n]}P_i$. 
Therefore,
\begin{equation}
    |\Delta_{\tau}|
        \le \sqrt{ \frac{d \sigma^2}{n}\sum_{t=0}^{T-1}\log\left(1 + \frac{M P_{\max} rn C_g \sum_{i\in\mathcal{I}^{(t)}}|h_i^{(t)}|^2}{d v^{(t)} \epsilon_g} \right) },
\end{equation}
which completes the proof by defining $C_g=4Q^2 G^2 (\Lambda + 1)( (1+\alpha L_{G})^2 + \Myfrac{\alpha^2 \sigma_H^2}{m_B})$.
}

{\color{black}
\subsection{Additional Experimental Results} \label{appendix:additional experiments}
To validate the generalization capacity of the proposed Air-meta-pFL, we conduct additional experiments on CIFAR10 \cite{krizhevsky2009learning} and Fashion-MNIST \cite{xiao2017fashion} data sets with the experimental parameters specified in Table~\ref{table:response experimental parameters}. 
The task of each device is to distinguish among $N=2$ different classes, which are pre-assigned to each device from 10 total classes, with $K=32$ shots for each class (see Fig. \ref{fig:experiment_illustration}).
Other parameters that are not listed here remain the same as in Sec.~\ref{subsec:Experimental Settings}. 
We use a simple CNN network that consists of two convolutional layers with ReLU activation and max pooling, followed by two fully connected layers for 10-class classification.

Fig.~\ref{fig:acc_vs_T_cifar10} and Fig.~\ref{fig:acc_vs_T_fmnist} show the test accuracy versus global rounds $T$ with Air-meta-pFL and other benchmarks on CIFAR10 [R7] and Fashion-MNIST [R8], respectively.
This result shows that the proposed Air-meta-pFL achieves comparable performance with vanilla meta-pFL \cite{fallah2020personalized}, and significantly outperforms FedAvg \cite{mcmahan2017communication} across different data sets, demonstrating its good generalization capabilities.

\begin{table}[t]
    \caption{Experimental parameters for CIFAR10 and Fashion-MNIST}
    \centering
    \begin{tabular}{llll}
        \hline
        \textbf{Parameter}             & \textbf{Value}                                                                                               & \textbf{Parameter}             & \textbf{Value}                                                          \\ \hline
 Mini-batch size                & $m_B=64$                                                                                                     & Data heterogeneity                        & $2$ classes of data for each device                                                                  \\ \hline
 Local SGD                      & $Q=5$   & Number of devices & $n=20$                                                                   \\ \hline
 Fraction of active device      & $r=0.5$ & Number of rounds        & $T=100$   \\ \hline
 Meta-learning rate             & $\eta=0.05$                                                                                        & Inner learning rate            & $\alpha=0.05$ \\ \hline
 Model & CNN          & Monte Carlo     & 10                                                          \\ \hline
        \end{tabular}   \label{table:response experimental parameters}
\end{table}

\begin{figure}[ht]
    \centering
    \subfigure[]{
        \centering
        \includegraphics[width=2.8in]{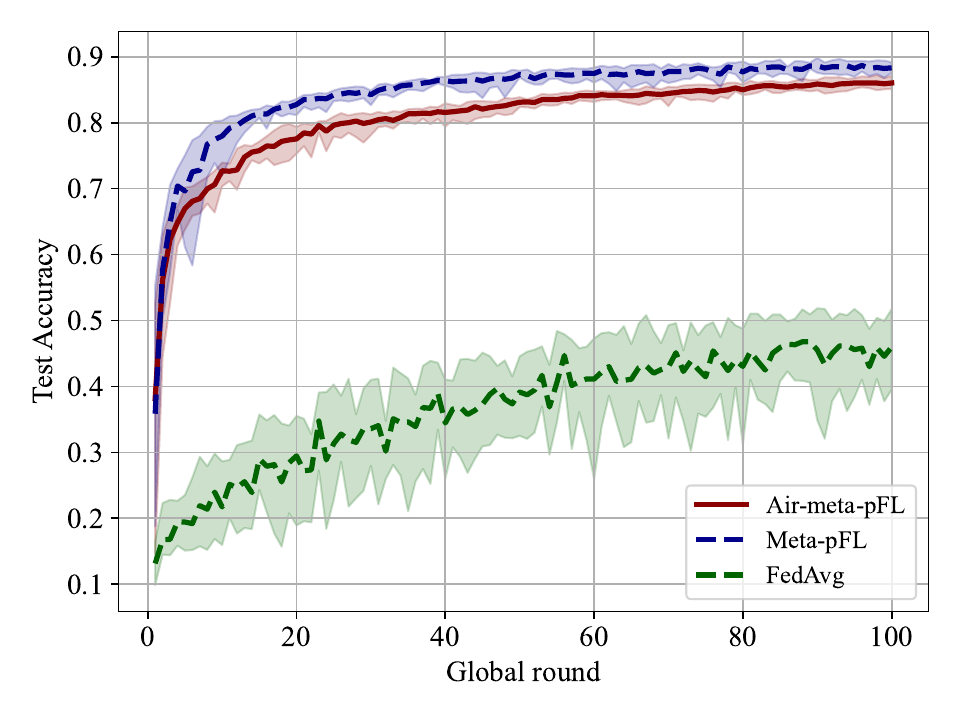}
        \label{fig:acc_vs_T_cifar10}
 }\hfill
    \subfigure[]{
        \centering
        \includegraphics[width=2.8in]{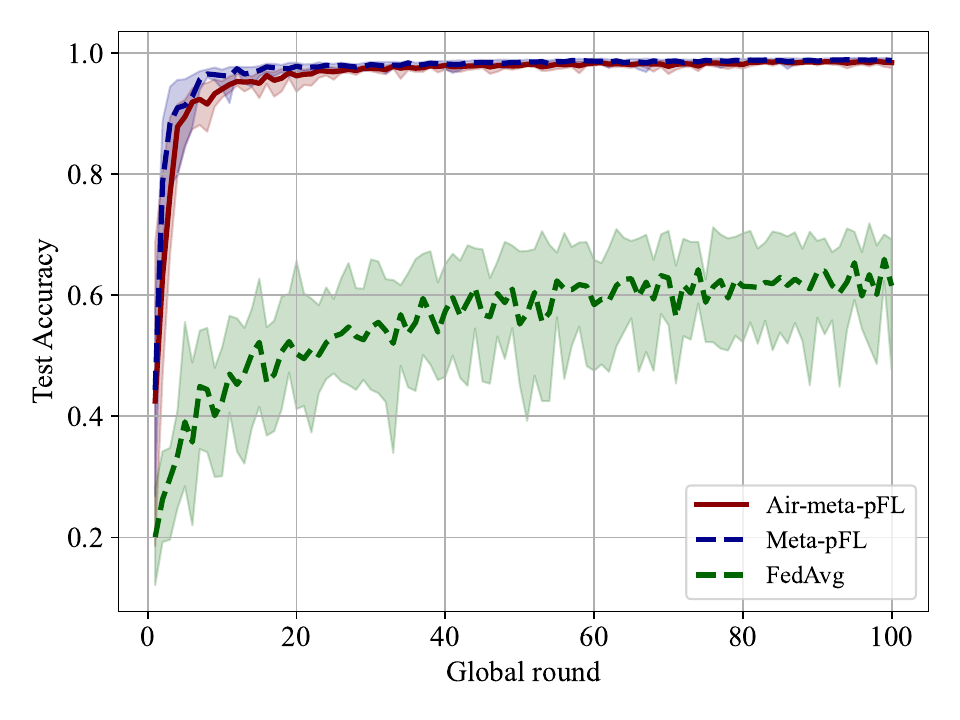}
        \label{fig:acc_vs_T_fmnist}
 }
    \caption{Test accuracy versus global rounds $T$ on data set (a) CIFAR10; and (b) Fashion-MNIST.}
    \label{fig:acc_vs_T_response}
\end{figure}
}

\bibliographystyle{IEEEtran}
\bibliography{ref}

\end{document}